\documentclass[
aip,
jcp,
reprint,
raggedbottom
]{revtex4-2}
\usepackage[T1]{fontenc}
\usepackage[utf8]{inputenc}

\usepackage{amsmath}
\usepackage{amssymb}
\usepackage{calc}
\usepackage{graphicx}
\usepackage{xcolor}
\usepackage{siunitx}\DeclareSIUnit\angstrom{\text{Å}}
\usepackage[hidelinks]{hyperref}
\usepackage{algorithm}
\usepackage{booktabs}
\usepackage{algpseudocode}
\usepackage{setspace}

\DeclareMathOperator{\E}{\mathbf{E}}

\newcommand*{\tran}{\mathsf{T}}

\begin{document}

\title{Using pretrained graph neural networks with token mixers as geometric featurizers for conformational dynamics}

\author{Zihan Pengmei}
\affiliation{Department of Chemistry and James Franck Institute, University of Chicago, Chicago, Illinois 60637, United States}
\author{Chatipat Lorpaiboon}
\affiliation{Department of Chemistry and James Franck Institute, University of Chicago, Chicago, Illinois 60637, United States}
\author{Spencer C. Guo}
\affiliation{Department of Chemistry and James Franck Institute, University of Chicago, Chicago, Illinois 60637, United States}
\author{Jonathan Weare}
\affiliation{Courant Institute of Mathematical Sciences, New York University, New York, New York 10012, United States}
\author{Aaron R. Dinner}
\email{dinner@uchicago.edu}
\affiliation{Department of Chemistry and James Franck Institute, University of Chicago, Chicago, Illinois 60637, United States}

\begin{abstract}
Identifying informative low-dimensional features that characterize dynamics in molecular simulations remains a challenge, often requiring extensive manual tuning and system-specific knowledge. Here, we introduce geom2vec, in which pretrained graph neural networks (GNNs) are used as universal geometric featurizers. By pretraining equivariant GNNs on a large dataset of molecular conformations with a self-supervised denoising objective, we obtain transferable structural representations that are useful for learning conformational dynamics without further fine-tuning. We show how the learned GNN representations can capture interpretable relationships between structural units (tokens) by combining them with expressive token mixers. Importantly, decoupling training the GNNs from training for downstream tasks enables analysis of larger molecular graphs (such as small proteins at all-atom resolution) with limited computational resources. In these ways, geom2vec eliminates the need for manual feature selection and increases the robustness of simulation analyses.
\end{abstract}

\maketitle

\section{Introduction}\label{sec:intro}

Molecular dynamics simulations can provide atomistic insight into complex reaction dynamics, but their high dimensionality makes them hard to interpret. 
Analyzing simulations thus relies on identifying low-dimensional representations (features), but care is needed in choosing them because they can strongly impact conclusions \cite{husic_optimized_2016, scherer_variational_2019, nagel_selecting_2023, arbon_markov_2024}.
Often features are selected manually, but doing so relies on system-specific intuition. Because developing intuition is often the goal of simulations, many researchers instead have turned to machine-learning methods for constructing features that capture observed variance (e.g., principal component analysis) \cite{tribello_using_2012, rohrdanz_determination_2011, boninsegna_investigating_2015} or decorrelate slowly according to the simulation data (e.g., the variational approach for Markov processes, VAMP) \cite{mardt_vampnets_2018, noe_variational_2013,chen_nonlinear_2019, wu_variational_2020, lorpaiboon_integrated_2020}. 
While these objectives are not always aligned with the reaction of interest \cite{trstanova2020local,strahan_long-time-scale_2021,nagel_selecting_2023,arbon_markov_2024}, such features can serve as useful intermediaries for computing reaction-specific statistics \cite{thiede_galerkin_2019,strahan_long-time-scale_2021} that provide a principled way for evaluating mechanisms \cite{bolhuis_reaction_2000, ma_automatic_2005,guo_dynamics_2024}.

Generally the inputs to the machine-learning methods above are internal coordinates such as distances between selected atoms and dihedral angles because they are invariant to translations and rotations of the system.  However, the nonlocal nature of these coordinates and/or their effects (e.g., the rotation of a dihedral angle in a polymer backbone) can make the resulting features both ineffective and hard to interpret, and these issues become more significant with system size\cite{ma_automatic_2005, bolhuis_reaction_2000}.  Additionally, it is not obvious how to represent permutationally invariant species such as solvent molecules with internal coordinates; recently introduced machine learning approaches for treating such species do not scale well with system size \cite{herringer2023permutationally}. 

Because atoms and their interactions (through bond or through space) can be viewed as the nodes and edges of graphs, molecular information can be readily encoded in graph representations (e.g., graph neural networks, GNNs).  Importantly, graph representations can be constructed in ways that respect the symmetries of molecular systems, with translational, rotational, and permutational invariance and equivariance.  Equivariance allows, for example, GNNs to output forces that rotate with the system, and appears to improve learning \cite{thomas2018tensor, anderson2019cormorant,batzner20223}. Owing to both their conceptual appeal and their performance, GNNs now dominate machine learning for force fields and molecular property prediction \cite{han2024survey, gilmer2017neural,battaglia_relational_2018,husic2020coarse,batzner20223}. They also are being successfully used to learn representations of larger molecules for tasks such as protein structure prediction, design, fold classification, and function prediction  \cite{jamasb_evaluating_2023, hermosilla_contrastive_2022, wang_learning_2023, zhang_protein_2023}.

The tasks above concern prediction of static properties, including structures.  Because molecular dynamics trajectories consist of sequences of structures, GNNs should be useful for identifying features for computing reaction statistics, and several groups have combined GNNs with VAMP \cite{mardt_vampnets_2018} to learn metastable states and relaxation time scales of both materials and biomolecules \cite{xie2019graph,soltani2022exploring,ghorbani2022graphvampnet, liu2023graphvampnets,huang2024revgraphvamp}. These groups report improved variational scores, convergence for shorter lag times, and more interpretable learned representations relative to VAMPnets based on fully connected networks. 
However, existing GNNs for analyzing dynamics do not readily scale to large numbers of atoms, so the graphs in these studies are small, either because the molecules are small, or only a subset of atoms (e.g., the C$_\alpha$ atoms of proteins) are used as inputs.
Furthermore, training these GNNs is computationally costly, limiting the number of architectures that can be explored and their use for other types of analyses.

The key idea of this paper is that GNNs can be pretrained using independent structural data prior to their use to analyze dynamics, thus decoupling GNN training from training for downstream tasks. Pretraining has transformed other domains such as natural language processing (NLP) and computer vision, enabling high-dimensional latent representations of words (``tokens'') \cite{vaswani2017attention} or images (``patches'') \cite{dosovitskiy2020image} to be learned by self-supervised, auto-regressive training. In NLP, word2vec pioneered the idea of using learnable vector representations for words by assigning them based on the word itself and its surrounding context \cite{mikolov2013distributed}; these representations could then be used for diverse downstream tasks. Inspired by word2vec, we propose geom2vec, an approach that leverages pretrained GNNs to learn transferable vector representations for molecular geometries.

Various pretraining strategies have been tried in molecular contexts \cite{guo2022self,hermosilla_contrastive_2022,zaidi2022pre,zhang_protein_2023,jamasb_evaluating_2023, chen2023structure}, but in contrast to complex pretraining and encoding schemes devised for specific classes of molecules \cite{zhang_protein_2023, jamasb_evaluating_2023}, we use a scheme that can be applied generally.  Building on the idea that corrupting data with noise and training a model to reconstruct the original data (denoising) can lead to learning meaningful representations for generative models \cite{sohl2015deep,ho_denoising_2020}, \citet{zaidi2022pre} showed that denoising atomic coordinates of structures of organic molecules significantly improved GNN performance on a number of molecular property prediction benchmarks.

Here, we show that this simple pretraining scheme also enables analysis of molecular dynamics simulations. Specifically, we pretrain GNNs using the same denoising objective and a dataset of structures of organic molecules obtained with density functional theory \cite{christensen2021orbnet} and analyze protein molecular dynamics simulations with the resulting representations.
We consider two tasks: learning slowly decorrelating modes with VAMP \cite{mardt_vampnets_2018} and identifying metastable states with the state predictive information bottleneck (SPIB) framework \cite{wang2021state, wang2024information}.
We show that neural networks trained using the GNN representations can readily take all non-hydrogen atoms in a small protein as inputs, enabling, for example, discovery of side chain dynamics that are important for folding.
By decoupling learning molecular representations from training for specific tasks, our method naturally accommodates alternative pretraining schemes \cite{jamasb_evaluating_2023, ni_pre-training_2024} and datasets \cite{vandermeersche_atlas_2024} (e.g., ones specific to particular classes of molecules), as well as other possible tasks \cite{hernandez_variational_2018, chen_discovering_2023, strahan_inexact_2023, jung_machine-guided_2023}.

\section{Methods}\label{sec:methods}

\begin{figure*}
    \centering
    \includegraphics{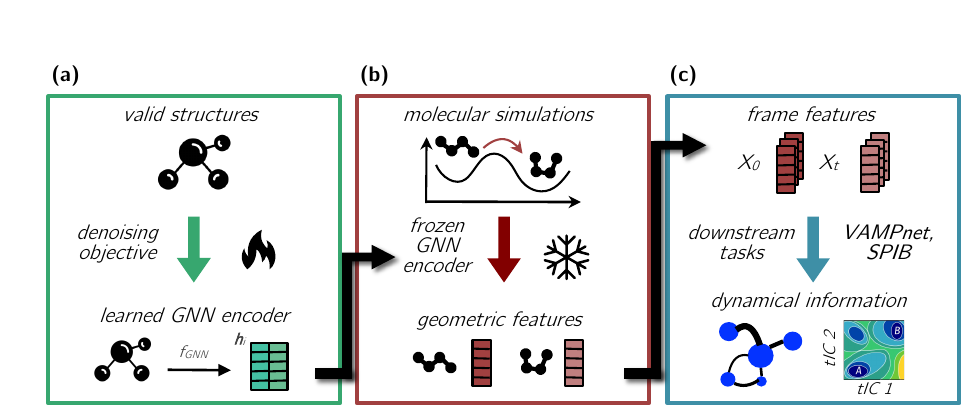}
    \caption{The geom2vec workflow. \textbf{(a)} A GNN encoder is pretrained using a denoising objective on a dataset of structures of diverse molecules. \textbf{(b)} Geometric representations for configurations from molecular simulations are obtained by performing inference with the pretrained GNN encoder. \textbf{(c)} The representations are used as inputs to a downstream task head (here, a VAMPnet or SPIB), which is trained separately.}
    \label{fig:workflow}
\end{figure*}

The basic idea of our method is to pretrain a GNN using a suitable task (here, denoising molecular coordinates) and then to use it with the resulting parameter values fixed as a feature encoder for other (downstream) tasks, as summarized in Figure \ref{fig:workflow}.   
In this section, we provide an overview of the network architecture that we use and then describe its elaboration for the pretraining and downstream tasks; further details are provided in the Appendices.
We refer to the workflow of transforming the atomic coordinates to representation vectors (i.e., features) and the use of those vectors as ``geom2vec.''

\subsection{Network architecture}

As noted above, our goal is learn a mapping from Cartesian coordinates to representation vectors via a GNN.
There are many existing GNN architectures from which to choose \cite{han2024survey}.  Here, we use the ViSNet \cite{wang2024enhancing} architecture, which is built on TorchMD-ET\cite{tholke2022torchmd}. These are both equivariant geometric graph transformers with modified attention mechanisms that suppress interactions between distant atoms; ViSNet goes beyond TorchMD-ET in using (standard and improper) dihedral angle information in its internal representations. We take the activations after the last message-passing layer as the representations for downstream computations.  These representations include three-dimensional vector features, which change appropriately with molecular translation and rotation, and one-dimensional scalar features, which are invariant to molecular translation and rotation. 
That is, 
\begin{equation}
f_\text{GNN}: \mathbb{R}^{3 N} \rightarrow \mathbb{R}^{(1+3) d N},
\end{equation}
where $N$ is the number of atoms, and $d$ is the number of features associated with each atom (equal to the dimension of the last update layer).
We represent the combined vector ($\mathbf{v}_i\in \mathbb{R}^{3 d}$) and scalar ($x_i \in \mathbb{R}^{d}$) features for atom $i$ by $\mathbf{h}_i \in \mathbb{R}^{(1+3) d}$.
We choose ViSNet because it was previously shown to give accurate molecular property predictions and accurate conformational distributions when used to learn a potential for molecular dynamics simulations. \cite{wang2024enhancing}  We briefly introduce the TorchMD-ET and ViSNet architectures in  Appendices~\ref{sec:torchet_arch} and \ref{sec:visnet_arch}, and we refer interested readers to the original publications and Ref.\ \citenum{pengmei2024pushinglimitsallatomgeometric} for further details.

\subsection{Pretraining by denoising}

For the pretraining, we draw random displacements from a multivariate Gaussian distribution and add them to the Cartesian coordinates of the molecules in the training set; we then train the GNN to predict the displacements. This process is designed to encourage the network to learn representations that capture the geometry of the molecular conformations, and it can be viewed as learning a force field with energy minima close to the training set geometries \cite{zaidi2022pre}.
We choose this objective because structural data are more readily available than energetic data, especially for large molecules such as proteins.

Here, we use the OrbNet Denali dataset, which consists of 215,000 molecules and complexes (with an average of 45 atoms) with 2.3 million conformations sampled from molecular trajectories \cite{christensen2021orbnet}.  We randomly select 10,000 conformations for validation and use the remainder for training.

Following \citet{zaidi2022pre}, we pretrain the model by passing the $(1+3)d$ features for each atom (graph node) from the ViSNet architecture through a gated equivariant block (Algorithm \ref{alg:gated}) that combines the $d$ scalar and vector features to obtain a three-dimensional vector that represents the predictions for the displacement of that atom.
We train for a fixed number of epochs and save the parameters resulting in the lowest mean squared error (MSE) between the predicted and added atom displacements for the validation set; for this comparison, we normalize the added atom displacements to have zero mean and unit variance.  Further details, including hyperparameters, are given in Appendix~\ref{appendix:training}. Depending on the network architecture, choice of hyperparameters, and graphics card available, the one-time pretraining can take from several hours to a few days.

\subsection{Use of the representations}

In this section, we discuss the operational details of using the pretrained GNN for downstream tasks in general (summarized in Figure \ref{fig:workflow}).  The specific downstream tasks that we use for our numerical demonstrations are described in Section \ref{sec:downstream}.

\subsubsection{Atom selection}

We first select the atoms of interest. For example, when analyzing simulations with explicit solvent, we may select only the solute coordinates.  
Given the molecular coordinates, the pretrained GNN yields a learned representation $\mathbf{h}_i$ for each selected atom $i$.
One can choose to base the calculations for the downstream tasks on the atomic representation vectors and/or the sums over their $d$ scalar and vector elements, but in most cases we reduce the size of the graph by coarse-graining it. 

\subsubsection{Coarse-graining}
Let $\mathcal{S} = \{S_1, \dots, S_M\}$ be a partition of the atoms into $M$ disjoint subsets, where each subset $S_m$ contains atoms belonging to a structural unit, such as a functional group or monomer in a polymer, depending on the system.
We pool the atomic representations for each $S_m$:
\begin{equation}
\bar{\mathbf{h}}_{S_m} = \sum_{i \in S_m} \mathbf{h}_i.
\end{equation}
Each resulting coarse-grained representation vector $\bar{\mathbf{h}}_{S_m}$ represents the geometric information of the atoms within its structural unit in an average sense. Following the NLP literature, we refer to the coarse-grained representations as structural ``tokens.'' 

How to partition the atoms is the user's choice, but many systems have a natural structure. For example, here we study proteins and pool the representations for the atoms in each amino acid. 
 In this case, the number of nodes in the graph is reduced by an order of magnitude.  Coarse-graining reduces the computational cost and memory, and it facilitates both learning long-range relationships and interpretation of the results (e.g., attention maps). 

\subsubsection{Feature combination}
To use the coarse-grained tokens in a learning task, we must combine their information in a useful fashion. How best to do so depends on the structural properties of interest, but we can generally lump approaches into two categories:
\begin{itemize}
    \item Pooling: We sum (or average) the coarse-grained tokens to obtain a single, global representation of the molecular system  $\bar{\mathbf{h}}\in \mathbb{R}^{(1+3)d}$. 
    Note that this approach is equivalent to pooling the atomic representation vectors $\mathbf{h}_i$.

    \item Token mixing: This approach applies a learnable mixing operation to the coarse-grained tokens to capture their interactions and dependencies \cite{carion2020end}. Token mixers allow for greater expressivity than direct pooling and can capture complex interactions between structural units.  
\end{itemize}
In this work, we employ two basic token mixers: (1) a standard transformer architecture \cite{vaswani2017attention}, which we refer to as ``SubFormer,''\cite{pengmei2023transformers} and  (2) an MLP-mixer architecture\cite{tolstikhin2021mlp}, which we refer to as ``SubMixer.'' 
Because the GNNs here employ a message passing architecture (Appendix~\ref{appendix:gnn}), the resulting features typically do not encode global geometric information. We show that this issue can be addressed by combining SubFormer and SubMixer with a special token that encodes global information \cite{pengmei2024technical} such as pairwise distances. Alternatively, they can be combined with geometric vector perceptrons (GVPs),  equivariant GNNs introduced for biomolecular modeling \cite{jing2020learning} to learn expressive positional encodings (see Algorithms \ref{alg:gvp} and \ref{alg:geom2vec} in Appendices~\ref{appendix:pseudo-code-gvp} and \ref{appendix:pseudo-code}).

\subsection{Output layers}

Ultimately, we combine the features from the different graph nodes and any graph-wide information (e.g., the CLS token of the transformer) and use an MLP to output quantities specific to a downstream task.  Here, we consider downstream tasks that require only scalar quantities, so we input only the scalar features to the MLP, but generally scalar (invariant) and vector (equivariant) quantities can be input and output.
We summarize the overall scheme in Algorithm~\ref{alg:geom2vec}.

\section{Downstream tasks}\label{sec:downstream}

To assess whether the geom2vec representations are useful for learning protein dynamics, we apply them to learning slowly decorrelating modes with VAMP \cite{mardt_vampnets_2018} and identifying metastable states with the state predictive information bottleneck (SPIB) framework \cite{wang2021state}.
As described previously, we apply the pretrained GNN to the coordinates from molecular dynamics trajectories and then use the resulting features as inputs to the desired task without further fine-tuning the GNN parameters.
In this section, we briefly describe the two downstream tasks mathematically.

\subsection{VAMPnets}\label{sec:vampnet}

Let $\mathbf{X}_t$ be a Markov process and define the correlation functions
\begin{align}
    C_{00} &= \E_{\mathbf{X}_0 \sim \mu}[\chi_0(\tilde{\mathbf{h}}(\mathbf{X}_0)) \chi_0^\tran(\tilde{\mathbf{h}}(\mathbf{X}_0))]  \\
    C_{0\tau} &= \E_{\mathbf{X}_0 \sim \mu}[\chi_0(\tilde{\mathbf{h}}(\mathbf{X}_0))\chi_\tau^\tran(\tilde{\mathbf{h}}(\mathbf{X}_\tau))]  \\
    C_{\tau\tau} &=\E_{\mathbf{X}_0 \sim \mu}[\chi_\tau(\tilde{\mathbf{h}}(\mathbf{X}_\tau)) \chi_\tau^\tran(\tilde{\mathbf{h}}(\mathbf{X}_\tau))],
\end{align}
where $\chi_0$ and $\chi_\tau$ are vectors of functions and the expectation is over trajectories initialized from an arbitrary distribution $\mu$.  In our case, $\mathbf{X}_t$ represents the molecular coordinates at time $t$ and $\tilde{\mathbf{h}}(\mathbf{X})$ are the molecular features from a pretrained GNN. 
The variational approach for Markov processes (VAMP)\cite{noe_variational_2013,mardt_vampnets_2018, chen_nonlinear_2019,wu_variational_2020, lorpaiboon_integrated_2020} states that $\chi_0$ and $\chi_\tau$ represent the slowest decorrelating modes (or collective variables; CVs) of the system when maximizing the VAMP-2 score
\begin{equation} \label{eq:vamp2}
    \operatorname{VAMP-2} = \left\lVert C_{00}^{-1/2} C_{0\tau} C_{\tau\tau}^{-1/2} \right\rVert_{\rm F}^2,
\end{equation}
where the subscript F denotes the Frobenius norm.

Operationally, the components of $\chi_0$ and $\chi_\tau$ are learned from data by representing them by parameterized functions (e.g., neural networks in VAMPnets \cite{mardt_vampnets_2018, chen_nonlinear_2019}; the output of geom2vec in our case) and maximizing \eqref{eq:vamp2}. 
VAMPnets require one to specify the output dimension $d_o$ \textit{a priori}. For the benchmark systems that we consider, we choose $d_o$ based on previous results in the literature. 

As discussed below (Section~\ref{sec:split}) we split each dataset into training and validation sets, evaluating the validation score every 10 training steps.
Each training step or validation step, we randomly draw a batch of trajectory-frame pairs spaced by $\tau$ to compute \eqref{eq:vamp2}.
We found that a large batch size (at least 1000 and usually 5000 for the examples here) was required to achieve a high validation score.
With smaller batch sizes, we encountered numerical instabilities when inverting the correlation matrices in the VAMP-2 loss function \eqref{eq:vamp2}.
A large batch size is also needed to minimize variance in the late phase of training because neural network outputs exhibit large changes between metastable states, where fewer trajectory-frame pairs contribute.
To prevent overfitting, we apply an early stopping criterion\cite{lorpaiboon_integrated_2020}: we stop training when the training VAMP score does not increase for 500 batches or the validation VAMP score does not increase for 10 batches. 
Further training details are given in Table~\ref{tab:hp_vamp_pretrain}.

\subsection{State Predictive Information Bottleneck (SPIB)}

In the information bottleneck (IB) framework, an encoder-decoder setup is used to learn a low-dimensional (latent) representation $\mathbf{z}$ that minimizes the information from a high-dimensional input $\mathbf{x}$ while maximizing the information about a target $\mathbf{y}$. The associated loss function is
\begin{equation}
\label{eq:ib}
\mathcal{L}_{\text{IB}} = I(\mathbf{z}, \mathbf{y}) - \beta I(\mathbf{x}, \mathbf{z}),
\end{equation}
where $I$ refers to the mutual information between two random variables:
\begin{equation}
    I(A, B) = \int p(A, B) \ln \frac{p(A, B)}{p(A)p(B)} \, dA \, dB,
\end{equation}
and the parameter $\beta$ controls the tradeoff between prediction accuracy and the complexity of the latent representation.

In the state predictive information bottleneck (SPIB) extension of IB \cite{wang2021state}, the inputs are molecular features at time $t$, $\tilde{\mathbf{h}}(\mathbf{X}_t)$, and the targets are state labels $s_t$ that indicate the state of the system at time $t$.
The latent representation and state labels are learned simultaneously by predicting the state labels at time $t+\tau$ given the molecular features at time $t$.

To learn the latent representation $\mathbf{z}$, SPIB maximizes the loss function
\begin{multline}
    \label{eq:spib}
    \mathcal{L}_\text{SPIB} = \E_{\mathbf{X}_0 \sim \mu, \mathbf{z} \sim p_\theta(\mathbf{z}|\tilde{\mathbf{h}}(\mathbf{X}_0))}
    \Biggl[
        \ln q_\theta(s_{\tau}|\mathbf{z})\\
        - \beta \ln \frac{p_\theta(\mathbf{z}|\tilde{\mathbf{h}}(\mathbf{X}_0))}{r_\theta(\mathbf{z})}
    \Biggr].
\end{multline}
The encoder generates the latent representation $\mathbf{z}$ from the input with probability $p_\theta(\mathbf{z}|\tilde{\mathbf{h}}(\mathbf{X})) = \mathcal{N}(\mathbf{z};\mu_\theta(\tilde{\mathbf{h}}(\mathbf{X})),\Sigma_\theta(\tilde{\mathbf{h}}(\mathbf{X})))$, which is a multivariate normal distribution with learned mean $\mu_\theta(\tilde{\mathbf{h}}(\mathbf{X}))$ and learned covariance $\Sigma_\theta(\tilde{\mathbf{h}}(\mathbf{X}))$. The decoder takes the latent representation $\mathbf{z}$ and returns the probability $q_\theta(s|\mathbf{z})$ of each state label $s$; $q_\theta(s|\mathbf{z})$ is represented by a neural network with output dimension equal to the number of possible state labels.
The quantity $r_\theta(\mathbf{z})$ is a prior.
The state labels are updated during training as follows:
\begin{equation}
    \label{eq:spib_refine}
    s_\tau = \operatorname{argmax}_s q_\theta(s|\mu_\theta(\tilde{\mathbf{h}}(\mathbf{X}_\tau))).
\end{equation}
It is important to note that \eqref{eq:spib_refine} allows the number of distinct states that are populated to fluctuate (unpopulated states are ignored).

We follow \citet{wang2021state} and use a variational mixture of posteriors for the prior \cite{tomczak2018vae}: 
\begin{equation}
    r_\theta(\mathbf{z}) = \frac{\sum_i \omega_i p_\theta(\mathbf{z|}u_i)} {\sum_i \omega_i},
\end{equation} 
where $\omega$ and $u$ are learned parameters.
In this work, we prepared the initial state labels by performing \textit{k}-means clustering on the CVs learned from VAMPnets based on distances between C$_\alpha$ atoms with $k=100$ clusters. Training reduces the number of distinct states that are populated to the estimated number of metastable states. Further training details are given in Table \ref{tab:hp_spib}. 

\section{Systems studied}
\label{sec:data}

We examine the performance of geom2vec for analyzing data from long molecular dynamics simulations of three well-characterized fast-folding proteins (chignolin, trp-cage, and villin) \cite{lindorff-larsen_how_2011}.
The data for each system is a single, unbiased simulation, which we assume approximately samples the equilibrium distribution. In this section, we introduce each system and briefly describe its structure and dynamics.

\subsection{Chignolin}\label{sec:sys_chignolin}
Chignolin is a 10-residue fast-folding protein with sequence YYDPETGTWY.
The folded state consists of three $\beta$-hairpin structures that are distinguished by hydrogen bonding between the threonine side chains and their dihedral angles, which interconvert on the nanosecond timescale \cite{bonati_deep_2021}. 
The trajectory that we analyze is 106~$\mu$s long at 340 K and saved every 0.2~ns  \cite{lindorff-larsen_how_2011}. For VAMPnet fitting, we choose $d_o = 3$.

\subsection{Trp-cage}\label{sec:sys_trpcage}

Trp-cage is a 20-residue fast-folding protein\cite{barua_trp-cage_2008}; here we study the K8A mutant with sequence DAYAQWLADGGPSSGRPPPS. 
Its secondary structure consists of an $\alpha$-helix (residues 2--9), a short $3_{10}$-helix (residues 11--14), and a polyproline II helix (residues 17--19); the protein takes its name from Trp6, which is in the core of the folded state. 
The trajectory is 208~$\mu$s long at 290 K and saved every 0.2~ns \cite{lindorff-larsen_how_2011}.
Previous studies of this trajectory generally identified the folded and unfolded states, with varying numbers of intermediates and misfolded states \cite{sidky2019high,wang2024information}.
For VAMPnet fitting, we choose $d_o = 4$.

\subsection{Villin}

The 35-residue villin headpiece subdomain (HP35) \cite{mcknight_thermostable_1996} is a fast-folding protein with sequence LSDEDFKA\-V\-FGMTRSAFANLPLWnLQQHLnLKEKGLF where nL refers to the unnatural amino acid norleucine. The K65nL/N68H/K70nL mutant was engineered to fold more rapidly \cite{kubelka_sub-microsecond_2006}.
The secondary structure of villin consists of three $\alpha$-helices at residues 3--10, 14--19, and 22--32.
Villin has a hydrophobic core centered on residues Phe6, Phe10, and Phe17.
The trajectory that we study is 125~$\mu$s long at 360~K and saved every 0.2~ns \cite{lindorff-larsen_how_2011}. Previous studies typically identified three states: a folded state, an unfolded state, and a misfolded state \cite{lorpaiboon_integrated_2020}. \citet{wang_novel_2019} proposed two primary folding pathways, where either the C-terminus or the N-terminus folds first, ultimately leading to the native state. Additionally, a cooperative hydrophobic interaction may facilitate a third folding pathway. For VAMPnet fitting, we choose $d_o = 3$.

\begin{figure*}[bt]
    \centering
    \includegraphics{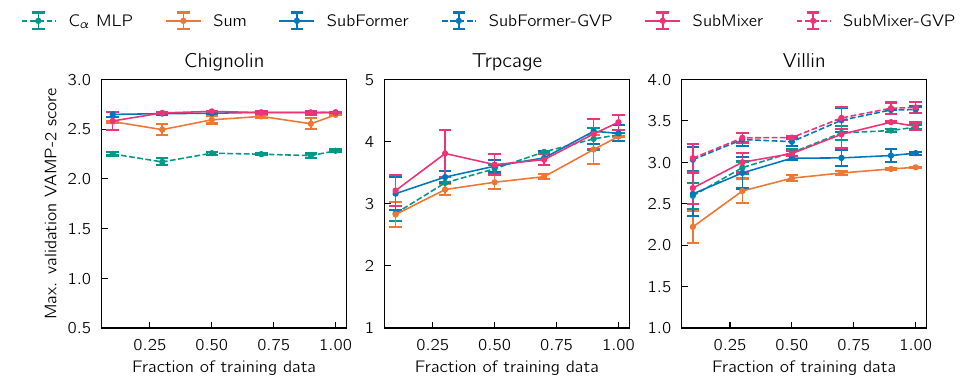}
    \caption{VAMPnets with various geom2vec architectures.  The amount of training data was varied by dividing the training data into 20 trajectory segments of equal length and then randomly selecting the indicated fraction for training. The validation set is held fixed as the second half of each trajectory. Error bars show standard errors over three independent runs.
    }
    \label{fig:vamp_data}
\end{figure*}

\subsection{Training-validation split}\label{sec:split}

Although previous studies used a random split\cite{mardt_vampnets_2018, ghorbani2022graphvampnet, chen_discovering_2023}, we observe that, due to the strong correlation between successive structures sampled by molecular dynamics simulations, a random split allows networks to achieve high validation scores even when they have memorized the training data rather than learned useful abstractions from it; the models then perform poorly on independent data. Consequently, for all our numerical experiments, we split the data into training and validation sets by time. That is, we select the first 50\% of the long trajectory for training and the remainder for validation. If we had access to multiple independent trajectories, randomly choosing trajectories for training and validation would also be appropriate. Some studies split the trajectory into equal segments and draw random segments for training and validation \cite{sidky2019high, wang2024information} (\textit{k}-fold cross-validation). When there are only two segments, this approach is identical to ours.  When every structure is its own segment, one recovers the random split.  Intermediate numbers of segments result in intermediate amounts of correlation between the training and validation sets. In cases where we vary the amount of training data, we first select the first 50\% of the trajectory and then divide only this half of the trajectory into segments that we draw randomly for training; the second half of trajectory is used as hold-out validation set. This approach is fundamentally different from cross-validation and minimizes the correlation between the training and validation datasets.

\section{Results}

\subsection{VAMPNets}
\label{sec:chignolinVAMP}
For each system and token mixer architecture pair that we consider, we independently train three VAMPnets using different random number generator seeds (and the training-validation split described in Section~\ref{sec:split}). We report the training and validation VAMP-2 scores for the different token mixer architectures for each of the three systems in Figures \ref{fig:train_curves} and \ref{fig:valid_curves}. For chignolin, GNNs with pooling (summing), SubMixer, and SubFormer reach approximately the same maximum validation score.  GNNs with SubMixer and SubFormer require fewer epochs to reach the convergence criteria, but they require more computational time per epoch, as we discuss in Section~\ref{sec:timing}. For both villin and trp-cage, the token mixers generally outperform pooling. 

\begin{figure}[bt]
    \centering
        \includegraphics{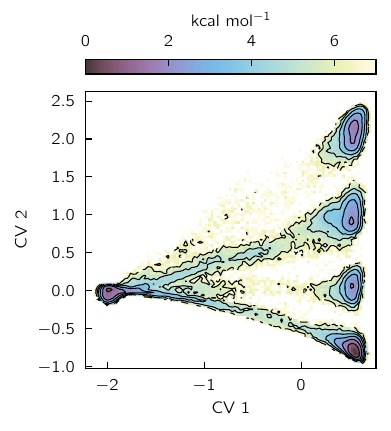}
    \caption{Potential of mean force (PMF) of chignolin as a function of the first two CVs learned by a VAMPnet trained with SubMixer. Contours are drawn every 1 kcal/mol. See Figure \ref{fig:vamp_chig_cvs}, \ref{fig:vamp_trpcage_cvs}, and \ref{fig:vamp_villin_cvs} for corresponding plots for other architectures and proteins.}
    \label{fig:vamp_chig_pmf}
\end{figure}

\begin{figure*}[bt]
    \centering
    \includegraphics[width=0.48\textwidth]{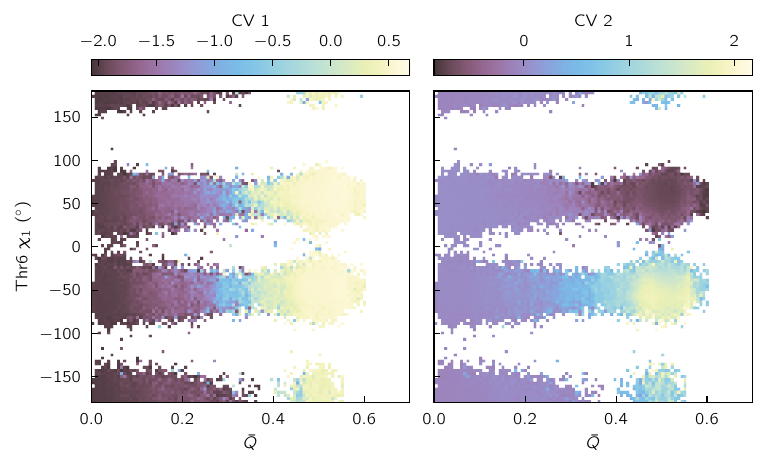}
    \includegraphics[width=0.48\textwidth]{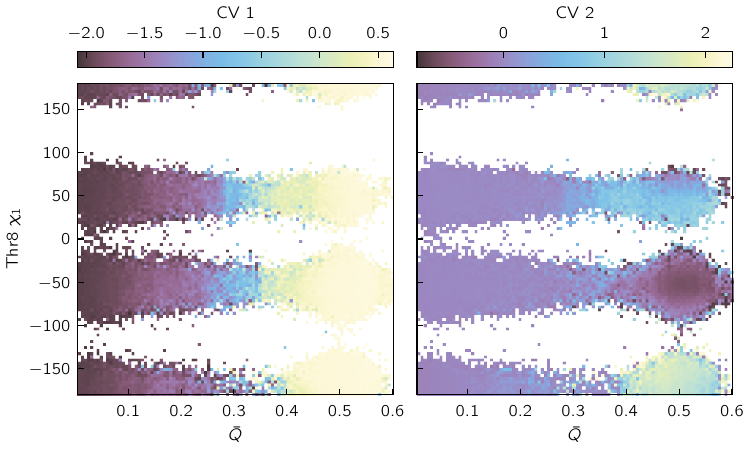}
    \caption{Chignolin VAMPnet (with SubMixer) CVs as a function of two physical coordinates: the fraction of native contacts and the $\chi_1$ side chain dihedral angle of Thr6 (left) or Thr8 (right). $\bar{Q}$ is the fraction of native contacts smoothed with a 1-ns moving window centered on each time point.  We define native contacts as two residues that are three or more positions apart in sequence and have at least one distance between non-hydrogen atoms that is less than 4.5 $\text{\AA}$ in the crystal structure (5AWL\cite{honda2008crystal}).   See Figures \ref{fig:vamp_trpcage_physical} and \ref{fig:vamp_villin_physical} for analogous plots for trp-cage and villin.
    }
    \label{fig:vamp_chig_physical}
\end{figure*}

Figure \ref{fig:vamp_data} displays the VAMP-2 scores for VAMPNets trained with different training dataset sizes, varying from 5\% to 50\% of the available data.
For chignolin, the GNNs clearly outperform a multilayer perceptron (MLP) that takes distances between pairs of C$_\alpha$ atoms as inputs; there is not a significant difference between pooling and the mixers considered.  For trp-cage and villin, the GNN with pooling consistently achieves the lowest scores.  The distance-based MLP and GNN with SubMixer perform comparably, presumably because the distances between pairs of C$_\alpha$ atoms are sufficient to describe the folding (in contrast to chignolin, as we discuss below). For villin, we combined SubMixer and SubFormer with GVP and augmented them with a global token; this enables the GNNs to outperform the distance-based MLP. The improvement is particularly striking for SubFormer. The models with GVP are more expressive because they use the equivariant features at the token mixing stage and directly mix them with global features after message-passing.
We note that the VAMP scores that we achieve are lower than published ones owing to our choice of the training-validation split (Section \ref{sec:split}) and output dimension $d_o$. While the amount of data does not significantly impact the performance for chignolin, the trp-cage and villin results suggest that additional data would permit achieving higher scores.

To visualize the results, we build histograms of learned CV-value pairs, which we convert to potentials of mean force (PMFs) by taking the negative logarithm (Figures~\ref{fig:vamp_chig_pmf}, \ref{fig:vamp_chig_cvs}, \ref{fig:vamp_trpcage_cvs}, and \ref{fig:vamp_villin_cvs}).  We also show average values of CVs as functions of physically-motivated coordinates (Figures~\ref{fig:vamp_chig_physical}, \ref{fig:vamp_trpcage_physical}, and \ref{fig:vamp_villin_physical}).

The advantages of the GNN architecture are well illustrated by the results for chignolin.
A previous machine-learning study of chignolin identified two slow CVs, one for the folding-unfolding transition and one distinguishing competing folded states\cite{bonati_deep_2021}. Our VAMPnets appear to recover these two CVs (Figures~\ref{fig:vamp_chig_pmf} and \ref{fig:vamp_chig_cvs}; results shown are with SubMixer), distinguishing folded and unfolded states with CV 1 and four folded states with CV 2. To understand the physical differences between the folded states, we plot CVs 1 and 2 as functions of the fraction of native contacts and the $\chi_1$ side chain dihedral of Thr6 or Thr8 (Figure~\ref{fig:vamp_chig_physical}). The fraction of native contacts clearly correlates with CV 1, consistent with earlier studies. CV 2 distinguishes the folded states by the configurations of Thr6 and Thr8 side chains, which can each occupy two rotamers, yielding four possible folded states. These side chain dynamics could not be detected by VAMPnets that take backbone internal coordinates as inputs, as is common, or even GNNs limited to backbone atoms (\citet{bonati_deep_2021} included distances to side chain atoms and then manually curated the inputs).
This makes clear the usefulness of the pre-training approach that we take here, which enables treating all the atoms with an architecture that supports both scalar and vector features.

\subsection{SPIB}\label{sec:spib_results}

\subsubsection{Trp-cage}
\begin{figure}[bt]
    \centering
    \includegraphics[width=\columnwidth]{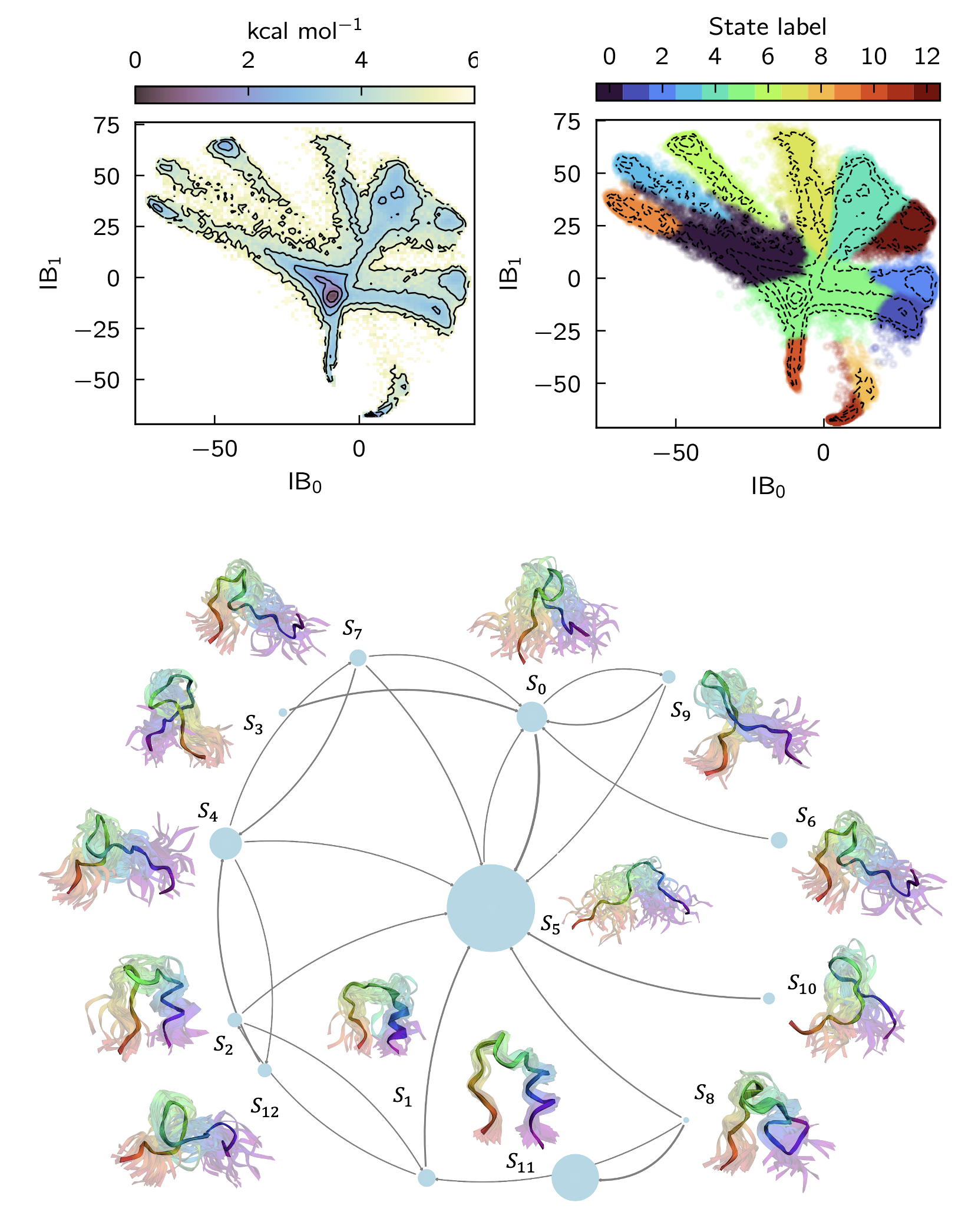}
    \caption{SPIB for trp-cage.  All results are obtained from a GNN with SubFormer-GVP token mixer.  (top left) PMF as a function of the first two information bottleneck coordinates (IBs). Contours are drawn every 1 kcal/mol. (top right) Same contours colored by SPIB assigned labels. (bottom) Learned Markov State Model.  The highlighted structures are chosen randomly from the trajectory. The N-terminus is violet and the C-terminus is red.}
    \label{fig:spib_network_trpcage_subformer}
\end{figure}

Figure \ref{fig:spib_network_trpcage_subformer} shows a Markov state model based on the states learned for trp-cage with a lag time of 20 ns. There are 13 metastable states. State $S_{11}$ represents the folded ensemble with a well-defined structure. In states $S_1$ and $S_2$ the $\alpha$-helix is folded, and the $3_{10}$-helix and polyproline II helix are unfolded. In contrast, in state $S_8$ the $3_{10}$-helix is folded, and the $\alpha$-helix and polyproline II helix are unfolded. State $S_5$ represents a fully unfolded state, which acts as a hub that connects all intermediate states and the folded state, $S_{11}$. States $S_3$, $S_4$, $S_6$, $S_7$, $S_9$, $S_{10}$, and $S_{12}$ are also largely unfolded and differ with regard to the specific conformations of the helices.

\begin{figure*}[hbt]
    \centering
    \includegraphics[width=\textwidth]{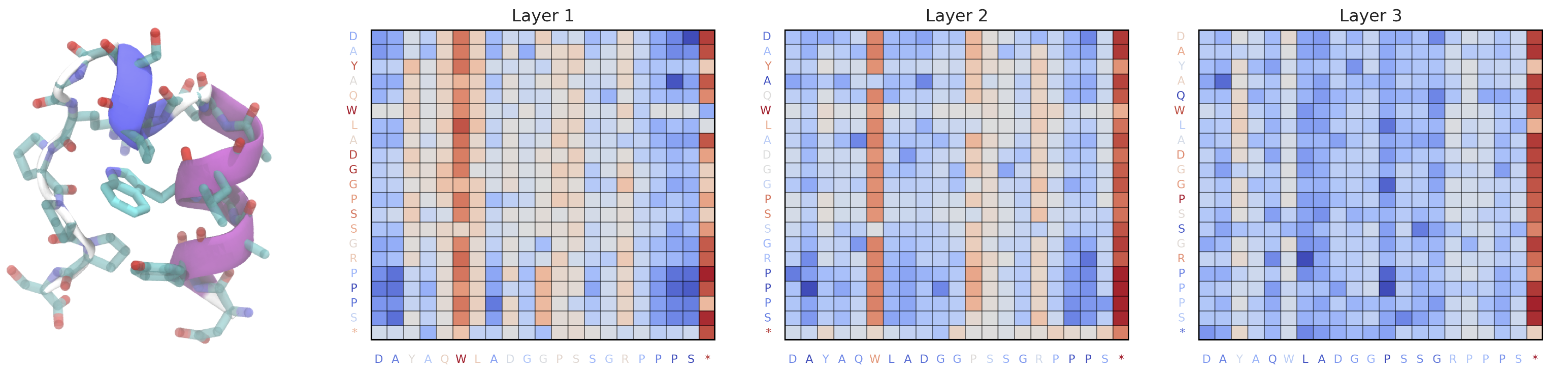}
    \caption{SPIB SubFormer-GVP attention. (left) A typical fully folded trp-cage structure (classified as $S_{11}$) with the central tryptophan residue (Trp6/W6) highlighted. (right three plots) Log-scaled averaged attention maps from three layers of the SubFormer block in the SubFormer-GVP architecture. The sequence is indicated by one-letter amino acid codes, and $*$ represents a global token that encodes pairwise distances between the C$_\alpha$ atoms. Tokens from query and key projections are colored according to the row-wise and column-wise sum of layer-wise attention weights. Results shown are for all structures in the trajectory. Results for individual states are in Figures \ref{fig:spib_attn_map_trpcage_0}, \ref{fig:spib_attn_map_trpcage_1}, and \ref{fig:spib_attn_map_trpcage_2}.}
    \label{fig:trpcage_attn_example}
\end{figure*}

\begin{figure}[htbp]
    \centering
    \includegraphics[width=\columnwidth]{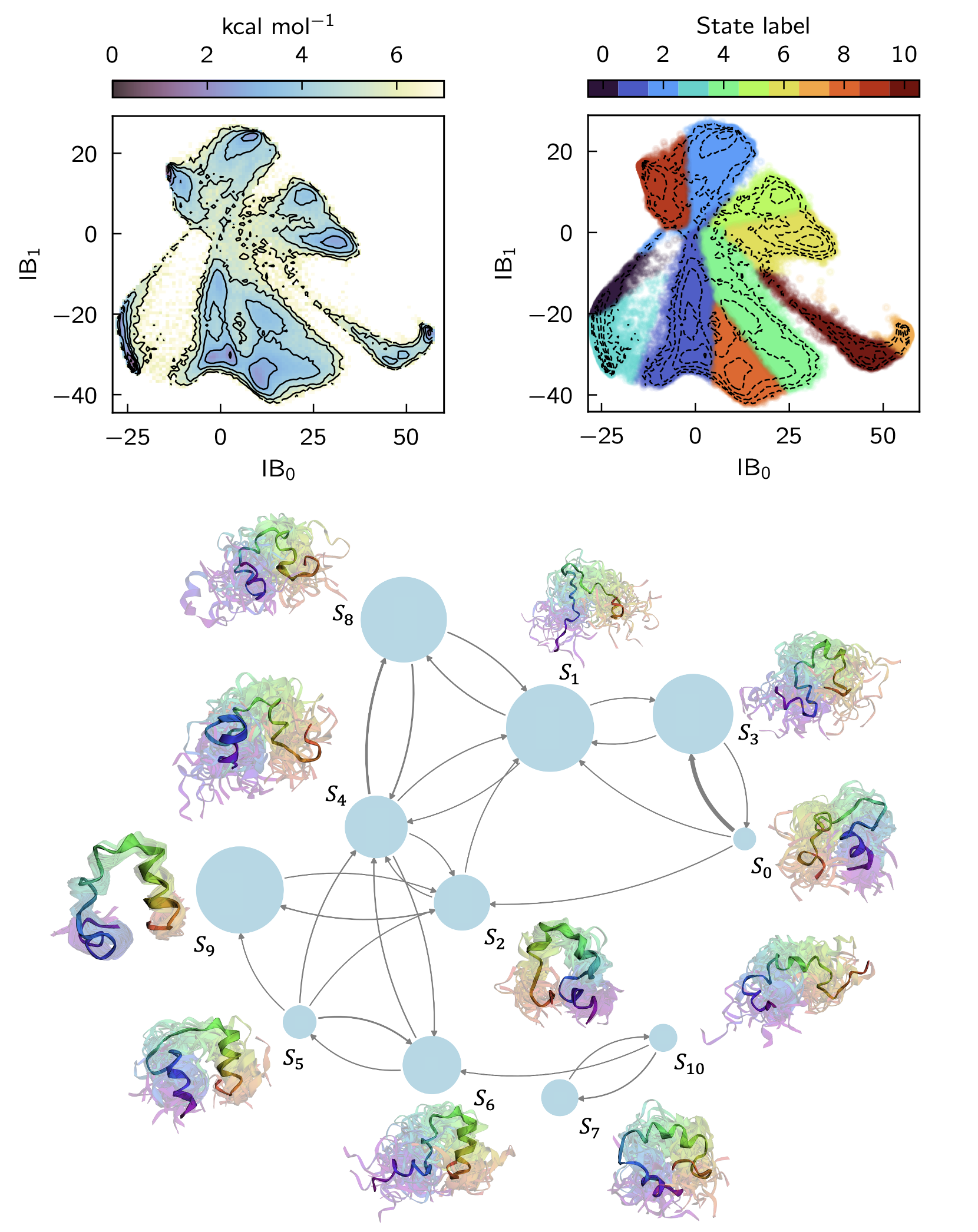}
    \caption{SPIB for villin.   All results are obtained from a GNN with SubFormer-GVP token mixer.  (top left) PMF as a function of the first two information bottleneck coordinates (IBs). Contours are drawn every 1 kcal/mol. (top right) Same contours colored by SPIB assigned labels. (bottom) Learned Markov State Model.  The highlighted structures are chosen randomly from the trajectory.  The N-terminus is violet and the C-terminus is red.}
    \label{fig:spib_network_villin_subformer}
\end{figure}
We show attention maps of the SubFormer blocks averaged over all structures in Figure \ref{fig:trpcage_attn_example} and over the structures in each state in Figures \ref{fig:spib_attn_map_trpcage_0}, \ref{fig:spib_attn_map_trpcage_1}, and \ref{fig:spib_attn_map_trpcage_2}. The attention maps are averaged over all heads of the transformer, yielding $(M+1) \times (M+1)$ matrices, where for trp-cage $M=20$ represents the number of sequence positions (tokens), and the additional row and column correspond to a global token that encodes pairwise distances between the C$_\alpha$ atoms \cite{pengmei2023transformers,pengmei2024technical}. The learned attention maps can be interpreted in terms of the structure. 
The residues that are most consistently activated across states and layers are Trp6 (W6) and Asp9--Ser14 (D9--S14), which roughly correspond to the 3$_{10}$-helix.  Tyr3 (Y3), Gln5 (Q5), Gly15 (G15), and Arg16 (R16) are activated in selected states.  The attention thus appears to track the packing of residues around Trp6.
Across all three layers, the global token remains highly active, consistent with the fact that the metastable states of trp-cage are reasonably well-characterized by the distances between the C$_\alpha$ atoms \cite{sidky2019high,strahan_long-time-scale_2021}. The fact that tokens other than the global token participate in the attention mechanism again underscores the ability of GNNs to go beyond distances between C$_\alpha$ atoms.

\subsubsection{Villin}

Figure \ref{fig:spib_network_villin_subformer} shows a Markov state model based on the states learned for villin HP35 with a lag time of 10 ns. There are 11 metastable states. State $S_9$ represents the fully folded structure, in which all three $\alpha$-helices are folded and packed compactly. States $S_1$ and $S_3$ correspond to fully unfolded states. 
In states $S_0$ and $S_2$ helix 1 is folded, suggesting a pathway in which folding initiates at the N-terminus, while in states $S_5$ and $S_6$ helix 3 is folded, suggesting a pathway in which folding initiates at the C-terminus.  Both of these pathways are discussed in the literature (see Ref.\ \citenum{wang_novel_2019} and references therein).

The attention maps (Figure \ref{fig:spib_attn_map_villin_0}, \ref{fig:spib_attn_map_villin_1}, and \ref{fig:spib_attn_map_villin_2}) exhibit patterns that correspond to features of the folded structure. Notably, the attention consistently focuses on the tokens representing Val9 (V9), Gly11 (G11), Met12 (M12), Arg14 (R14), Pro21 (P21), and Trp23 (W23). 
These residues correspond roughly to the turns between helices.  In the attention maps for states 0, 2, 5, and 6, the tokens corresponding to helix 3 feature prominently; the tokens corresponding to helix 1 are also activated in state 0.  The attention maps thus suggest that the network tracks the folding and packing of the helices.

It is interesting to compare our attention maps with those of \citet{ghorbani2022graphvampnet} and \citet{huang2024revgraphvamp}. Those studies leverage a graph attention network (GAT) \cite{velivckovic2017graph} to enhance expressive power and interpretability of their models. GAT computes representations of each node by attending to its one-hop neighboring nodes, which captures local dependencies but fails to model long-range interactions. In contrast, GNN-Transformer hybrids such as SubFormer \cite{pengmei2023transformers} allocate short-range interactions to the MP-GNN and use the self-attention mechanism for long-range interactions. This approach not only supports multimodal features (e.g., a global token) but also enables distant nodes to attend to each other, regardless of graph distance. This difference is evident in the attention map patterns: GAT attention maps \cite{ghorbani2022graphvampnet, huang2024revgraphvamp} show predominantly diagonal patterns, reflecting a focus on local neighborhoods, while SubFormer-GVP attention maps reveal vertical, blockwise, and global patterns, reflecting instead a focus on specific amino acids and their long-range interactions.

\section{Computational requirements}
\label{sec:timing}

Equivariant geometric GNNs use both invariant and equivariant features to capture the three-dimensional structure of molecules. For typical numbers of features, the memory and time requirements are expensive even for small graphs. To illustrate, we show the memory and time requirements for inference using a TorchMD-ET GNN with a small batch size with varying numbers of hidden channels (numbers of features) for trp-cage (144 non-hydrogen atoms) and villin (272 non-hydrogen atoms) in Figure \ref{fig:combined_usage}. Even this already requires tens of gigabytes of memory and several seconds; a more complex architecture like ViSNet is expected to increase the memory and time requirements by roughly 50\%.  The memory and time scale linearly with both the number of atoms in the graph and the batch size.  We use a batch size of 5000 for VAMP (except for GVP variants, for which we use 1000) and 1000 for SPIB, making training a GNN on the fly prohibitive, as we discuss in further detail below.

Geom2vec decouples training the GNNs and the networks for the downstream tasks.  This allows us to use a small batch size for pretraining the GNNs (which need be done only once), and the networks for the downstream tasks take as inputs the tokens, which are fewer in number than the number of graph nodes.  For example, here, the number of graph nodes is the number of non-hydrogen atoms, while the number of tokens is the number of amino acids, which is an order of magnitude smaller.  

\begin{figure*}[htb]
    \centering
    \includegraphics{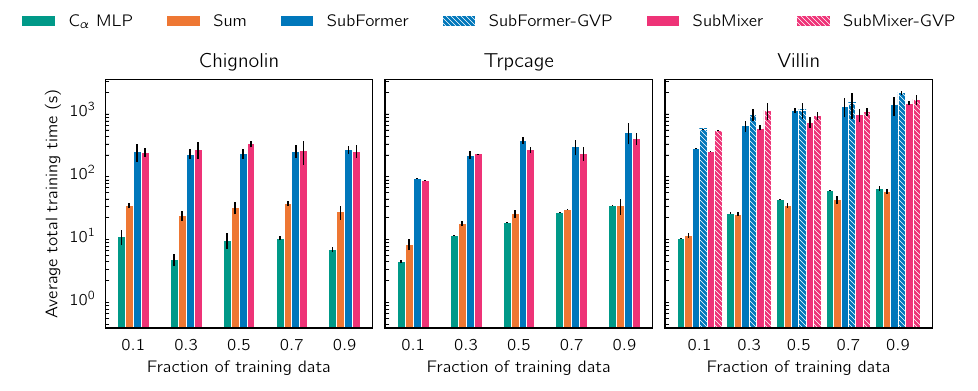}
    \caption{Computational time for training a single VAMPNet. We employ early stopping and stop training when the training VAMP score does not increase for 1000 batches or the validation VAMP score does not increase for 10 batches.  Times reported are averages over three training runs, with validation performed at each step using the second half of each trajectory, as depicted in Figure \ref{fig:valid_curves}. All times are for training on a single NVIDIA A40 GPU.}
    \label{fig:train_timing}
\end{figure*}

The computational costs for training VAMPnets with different token mixers are shown in Figure \ref{fig:train_timing}. The simplest GNN using pooling is not much more computationally costly than an MLP that takes distances between C$_\alpha$ atoms as inputs.  The GNNs with token mixers are about an order of magnitude more computationally costly but still manageable (hundreds of seconds) even without advanced acceleration techniques such as flash-attention or compilation. We expect the memory and computational requirements to scale with token number quadratically for SubFormer and subquadratically for SubMixer (depending on the expansion dimension in the token-mixing blocks); these requirements should scale linearly with respect to embedding dimension and network depth. 

To estimate the memory usage and training time for a VAMPnet based on a GNN without pretraining, we consider a TorchMD-ET model with specific settings: a batch size of 1000, a hidden dimension of 64, and 6 layers; we assume 50 epochs are required to converge. For trp-cage, a batch size of 100 required 10.64~GB of memory. Since memory usage and training time should scale linearly with batch size, we estimate that increasing the batch size to 1000 would raise the memory usage to approximately $10.64~\text{GB} \times 10 = 106.4~\text{GB}$. Similarly, we estimate the training time at this larger batch size to be around 1.8 hours (excluding validation). The corresponding numbers for villin, which is about twice the size, are proportionally larger (219.0~\text{GB} and 4.0 hours for training). As noted above, employing a more complex architecture like ViSNet would further increase both time and memory requirements by roughly 50\%. In the present study, we employ a batch size of 5000 for VAMP (except for GVP variants, as noted above); the batch size for SPIB is 1000, but it generally requires more iterations to converge.  While rough, these estimates show that without pretraining equivariant GNNs for analyzing molecular dynamics are beyond the resources available to most researchers.  By contrast, with pretraining, they are well within reach (Figure \ref{fig:train_timing}).

\section{Conclusions}

In this paper, we use pretrained GNNs to convert molecular conformations into rich vector representations that can then be used for diverse downstream tasks. Decoupling the training of the GNNs and the networks for the downstream tasks dramatically decreases the memory and computational time requirements.  Here, we focused on downstream tasks concerned with analyzing dynamics in molecular simulations, specifically VAMP and SPIB.  For these tasks, we were able to use equivariant GNNs that take all non-hydrogen atoms of small proteins as inputs for the first time.  The results for folding and unfolding of small proteins show that the GNNs use information beyond distances between C$_\alpha$ atoms, which are commonly used as input features.

For pretraining, we used a simple denoising task with a dataset of structures of diverse molecules.  Given that this dataset was not specific to proteins and/or VAMP and SPIB, we expect the present models to generalize to other classes of molecules and tasks, but quantitative tests on a wider variety of systems and tasks remain to be done. It would be interesting to investigate whether more complex pretraining strategies \cite{ni_pre-training_2024, liao_generalizing_2024} together with datasets specific to the class of molecules of interest (e.g., those specific to proteins\cite{sillitoe_cath_2021, vandermeersche_atlas_2024, barrio-hernandez_clustering_2023}) can improve performance.  We also explored a variety of token mixers and observed that more complex architectures were able to yield better results, which suggests there is scope for further engineering in this regard; these should be informed by ablation studies.  


In our tests, we took care to split the dataset in a way that minimized the correlation between the training and validation datasets, and we believe that this should be standard practice.  Because the data consisted of long, unbiased trajectories \cite{lindorff-larsen_how_2011}, there were relatively few events of interest (here, folding and unfolding).  Adapting our approach to methods that take short trajectories \cite{strahan_predicting_2023,strahan_inexact_2023}, which allow for greater control of sampling \cite{strahan_predicting_2023,strahan2024bad}, is an important area of study for the future.

\section*{Acknowledgments}
We thank D. E. Shaw Research for making the molecular dynamics trajectories available to us.  This work was supported by National Institutes of Health award R35 GM136381 and National Science Foundation award DMS-2054306.
S.C.G. acknowledges support from the National Science Foundation Graduate Research Fellowship under Grant No. 2140001. Z.P. was supported with funding from the University of Chicago Data Science Institute's AI+Science Research Initiative. This work was completed with computational resources administered by the University of Chicago Research Computing Center, including Beagle-3, a shared GPU cluster for biomolecular sciences supported by the NIH under the High-End Instrumentation (HEI) grant program award 1S10OD028655-0.

\section*{Data availability}
Data sharing is not applicable to this article as no new data were created or analyzed in this study.
Code for our implementation and examples are available at \url{https://github.com/dinner-group/geom2vec}.
\section*{References}
%

\clearpage
\onecolumngrid
\appendix

\renewcommand{\thesection}{\Alph{section}}
\renewcommand{\thefigure}{S\arabic{figure}}
\renewcommand{\theequation}{\thesection\arabic{equation}}
\renewcommand{\thetable}{S\arabic{table}}

\setcounter{subsection}{0}
\setcounter{section}{0}
\setcounter{equation}{0}
\setcounter{table}{0}
\setcounter{figure}{0}

\section{GNN architectures}\label{appendix:gnn}

\subsection{Equivariant GNNs}

Assuming $\mathbf{R} = (\mathbf{r}_1, \mathbf{r}_2, \ldots, \mathbf{r}_N) \in \mathbb{R}^{3N}$ defines the molecular conformation, where $\mathbf{r}_i \in \mathbb{R}^3$ represents the three-dimensional (3D) coordinates of the $i$-th atom, two of the most important symmetries for a GNN to obey are translation and rigid-body rotation invariances. These invariances require that the output of the GNN, $f(\mathbf{R})$, remains unchanged when the molecule is translated by a vector $\mathbf{t} \in \mathbb{R}^3$, i.e., $f(\mathbf{R}) = f(\mathbf{R} + \mathbf{t})$, or when the molecule undergoes a rigid body rotation represented by a rotation matrix $P \in SO(3)$, i.e., $f(\mathbf{R}) = f(P\mathbf{R})$. Translation invariance can be easily realized by using the relative displacements or internal coordinates of atoms. On the other hand, rotational invariance and equivariance\cite{} can be realized in two ways:
\begin{enumerate}
    \item representing the relative atom positions by internal coordinates (e.g., bond distances, bond angles, and dihedral angles) as in conventional classical force fields (``scalar-based network'') \cite{schutt2018schnet, gasteiger2020directional,tholke2022torchmd, wang2024enhancing};
    \item encoding the relative displacement vectors with spherical harmonics, which are then propagated with tensor products (``group-equivariant network'') \cite{thomas2018tensor, anderson2019cormorant,batzner20223}.
\end{enumerate}
The two approaches are theoretically and practically equivalent\cite{villar2021scalars, tholke2022torchmd,wang2024enhancing}, but group-equivariant networks need to store the features for each irreducible representation and are more computationally costly owing to the tensor products.
Therefore, we chose to employ scalar-based GNNs. 

\subsection{TorchMD-ET architecture}
\label{sec:torchet_arch}

The TorchMD-ET architecture \cite{tholke2022torchmd} is structured around three components that process and encode molecular geometric information effectively.
\begin{enumerate}
    \item \textbf{Embedding layer}: encodes atomic types and interatomic distances using exponential radial basis functions (eRBFs). Each distance $d_{ij}$ within a cutoff $r_{\text{cut}}$ is transformed by
    \begin{equation}
    \operatorname{eRBF}_k(d_{ij}) = \phi(d_{ij}) \exp\left(-\beta_k(\exp(-d_{ij}) - \mu_k)^2\right),
    \end{equation}
    where $\phi(d_{ij})$ is a cosine cutoff function ensuring smooth transitions to zero beyond $r_{\text{cut}}$.

    \item \textbf{Modified attention mechanism}: incorporates edge information through an extended dot-product attention mechanism that integrates interatomic distances with distance kernels $DK$ derived from eRBFs ($\odot$ denotes element-wise multiplication):
    \begin{equation}
    A = \operatorname{SiLU}\left(\sum_{k=1}^{F} Q_k \odot K_k \odot DK_k\right) \cdot \phi(d_{ij}).
    \end{equation}
    The attention mechanism weights are then used to compute scalar features and filters $q_{\alpha i}, s_{\alpha ij} \in\mathbb{R}^F$ ($\alpha =1, 2, 3$) for the update layer:
    \begin{align}
        s_{1ij}, s_{2ij}, s_{3ij} &= V_j \odot DV \\
        q_{1i}, q_{2i}, q_{3i} &= W \left(\sum_{j\in \mathcal{N}(i)\setminus i} A_{ij} \cdot s_{3ij}\right)
    \end{align}
    where $V_j$ and $DV$ are attention value and distance projections analogous to $Q$, $K$, and $DK$.

    \item \textbf{Update layer}: updates both scalar ($\Delta x_i$) and vector ($\Delta v_i$) features using $q_{\alpha i}$ and $s_{\alpha ij}$:
    \begin{align}
        \Delta x_i &= q_{1i} + q_{2i} \odot ( U_1 \mathbf{v}_i\cdot U_2 \mathbf{v}_i)\\
        \Delta \mathbf{v}_i &= q_{3i}\cdot U_3\mathbf{v}_i + \sum_{j} \left(s_{1ij} \odot \mathbf{v}_j + s_{2ij} \odot \frac{\mathbf{r}_i - \mathbf{r}_j}{\|\mathbf{r}_i - \mathbf{r}_j\|}\right).
    \end{align}
    This combines scalar and directional vector information describing the local atomic environment.
\end{enumerate}

\subsection{ViSNet}\label{sec:visnet_arch}

ViSNet\cite{wang2024enhancing} builds on TorchMD-ET by incorporating additional scalar features into its architecture. These additional features include information about angles, dihedral angles, and improper dihedral angles. This information is computed in an efficient fashion by first associating with each atom information about its neighbors:
\begin{align}
    \mathbf{u}_{ij} &= \mathbf{r}_{ij}/\|\mathbf{r}_{ij}\|, \\
    \mathbf{v}_{i} &= \sum_{j\in\mathcal{N}(i)\setminus i} \mathbf{u}_{ij}, \\
    \mathbf{w}_{ij} &= \mathbf{v}_i - ( \mathbf{v}_i \cdot \mathbf{u}_{ij} ) \mathbf{u}_{ij},
\end{align}
where  $\mathbf{u}_{ij}$ is the unit vector pointing from atom $i$ to neighboring atom $j$, and $\mathbf{v}_i$ is the sum of all such unit vectors around atom $i$. Then the inner product
\begin{equation}
    \mathbf{w}_{ij} \cdot \mathbf{w}_{ji} = \sum_{m\in\mathcal{N}(i)\setminus i}\sum_{n \in \mathcal{N}(j)\setminus j}\cos(\varphi_{mijn})
\end{equation}
 represents the dihedral angle ($\varphi_{mijn}$) information around inner atoms $i$ and $j$, with neighbors indexed by $m$ and $n$, respectively.

\subsection{Gated equivariant block}\label{sec:gated_equiv_block}
The output layer of the networks consists of gated equivariant blocks\cite{weiler_3d_2018, schutt_equivariant_2021}, each of which combines the scalar ($x_i$) and vector ($\mathbf{v}_i$) features from graph node $i$ of the previous layer (Algorithm \ref{alg:gated}). 

\begin{algorithm}[H]
\setstretch{1.25}
\caption{Gated equivariant block (GEB)}
\label{alg:gated}
\begin{algorithmic}[1]
\Require $x_i \in \mathbb{R}^{d_i}$, $\mathbf{v}_i\in \mathbb{R}^{3 d_i}$
\State $v_i^1, \mathbf{v}_i^2\gets \| W\mathbf{v}_i\|, W\mathbf{v}_i$ \Comment{$v_i^1\in \mathbb{R}^{d_i}$, $\mathbf{v}_i^2\in \mathbb{R}^{3 d_o}$}
\State $x_i\gets \operatorname{Concat}(x_i, v_i^1)$ \Comment{$x_i\in \mathbb{R}^{2d_i}$}
\State $x_i\gets W(\operatorname{SiLU}((W x_i + b))) + b$ \Comment{$x_i\in\mathbb{R}^{2d_o}$}
\State $[x_i, g_i] \gets x_i$ \Comment{split $x_i$ into $x_i, g_i \in\mathbb{R}^{d_o}$}
\State $x_i \gets \operatorname{SiLU}(x_i)$
\State $\mathbf{v}_i\gets g_i\odot \mathbf{v}_i^2$ \Comment{scales vector features, maintains equivariance}
\State \Return $x_i$, $\mathbf{v}_i$ \Comment{$x_i\in \mathbb{R}^{d_o}$, $\mathbf{v}_i\in \mathbb{R}^{3 d_o}$}
\end{algorithmic}
\end{algorithm}

\newpage
\section{Geometric vector perceptron}
\label{appendix:pseudo-code-gvp}

Geometric vector perceptron (GVP) was one of the first architectures to incorporate vector features for protein modeling \cite{jing2020learning}. Here, we use the GVP architecture as an equivariant token mixer to combine scalar and vector features (Algorithm~\ref{alg:gvp}).

\begin{algorithm}[H]
\setstretch{1.25}
\caption{Geometric vector perceptron (GVP)}
\label{alg:gvp}
\begin{algorithmic}[1]
\Require $x_i \in \mathbb{R}^{d_i}$, $\mathbf{v}_i\in \mathbb{R}^{3 d_i}$

\State $\mathbf{v}_h \gets W_h \mathbf{v}_i$ \Comment{project vector features, $\mathbf{v}_i\in\mathbb{R}^{3 d_h}$}
\State $x_h \gets \|\mathbf{v}_h\|_2$ \Comment{compute the (row-wise) norm of projected vectors, $x_h \in \mathbb{R}^{d_h}$}
\State $x_{h+d_i} \gets \operatorname{Concat}(x_h, x_i)$ \Comment{ $x_{h+d_i} \in \mathbb{R}^{d_h + d_i}$}
\State $x_m \gets W_m x_{h+d_i} + b_m$ \Comment{$x_m \in \mathbb{R}^{d_o}$}
\State $x'_i \gets \sigma(x_m)$ \Comment{$x'_i \in \mathbb{R}^{d_o}$}

\State $\mathbf{v}_o \gets W_o \mathbf{v}_h$ \Comment{project vector features to $\mathbb{R}^{3 d_o}$}

\State $\mathbf{v}'_i \gets \sigma(\|\mathbf{v}_o\|_2) \odot \mathbf{v}_o$ \Comment{column-wise multiplication with vector gate (on row-wise norm), $\mathbf{v}'_i \in \mathbb{R}^{3 d_o}$}

\State \Return $x'_i, \mathbf{v}'_i$

\end{algorithmic}
\end{algorithm}

\clearpage
\newpage
\section{Pseudo-code of overall architecture} 
\label{appendix:pseudo-code}
\begin{figure*}[!h]
\begin{minipage}{\linewidth}
\begin{algorithm}[H]
\setstretch{1.25}
\caption{geom2vec}
\label{alg:geom2vec}
\begin{algorithmic}[1]
\Require Atomic numbers $z_i$, positions $\mathbf{r}_i\in \mathbb{R}^{3}$ ($i = 1,\dots,N$), coarse-grained mapping $\mathcal{S} = \{S_1, \dots, S_M\}$, mixer method \{None, SubFormer, SubMixer, SubGVP\}, global features $\mathbf{G}\in \mathbb{R}^{d_g}$
\State $x_i^{\text{MP}}, \mathbf{v}_i^{\text{MP}}\gets \operatorname{GNN}(z_i,\mathbf{r}_i)$ \Comment{atomistic featurization, $x_i^{\text{MP}}\in \mathbb{R}^{d_k}$, $\mathbf{v}_i^{\text{MP}}\in \mathbb{R}^{3 d_k}$}

\State $x_m^{\text{MP}}, \mathbf{v}_m^{\text{MP}} \gets \sum_{i \in S_m} x_i^{\text{MP}},\sum_{i \in S_m} \mathbf{v}_i^{\text{MP}} $ \Comment{coarse-graining with mapping $\mathcal{S}$, $x_m^{\text{MP}} \in \mathbb{R}^{ d_k}$,$\mathbf{v}_m^{\text{MP}} \in \mathbb{R}^{3 d_k}$}

\If {None}
    \State $x^{\text{MP}}, \mathbf{v}^{\text{MP}}\gets \sum_{m} x_m^{\text{MP}},\sum_{m} \mathbf{v}_m^{\text{MP}}$ \Comment{sum over indices $m$ for direct pooling, $x^{\text{MP}} \in \mathbb{R}^{ d_k}$,$\mathbf{v}^{\text{MP}} \in \mathbb{R}^{3 d_k}$}
    \State $x, \mathbf{v} \gets \operatorname{GEB}(x^{\text{MP}}, \mathbf{v}^{\text{MP}})$ \Comment{Algorithm~\ref{alg:gated}}
\ElsIf {SubFormer}
    \State $x_m, \mathbf{v}_m \gets \operatorname{GEB}(x_m^{\text{MP}}, \mathbf{v}_m^{\text{MP}})$ \Comment{apply GEB to input features}
    \If{global token}
        \State $g \gets \operatorname{MLP}(\mathbf{G})$ \Comment{project global features $\mathbf{G}$ to $\mathbb{R}^{d_k}$}
        \State $x \gets \operatorname{TransformerEncoder}([x_{m}, g])$ \Comment{pass regular and global tokens through the transformer encoder}
    \Else
        \State $x \gets \operatorname{TransformerEncoder}(x_m)$ \Comment{pass only the regular tokens through the transformer encoder}
    \EndIf

\ElsIf{SubMixer}
    \State $x_m, \mathbf{v}_m \gets \operatorname{GEB}(x_m^{\text{MP}}, \mathbf{v}_m^{\text{MP}})$ \Comment{apply GEB to input features}
    \If{global token}
        \State $g \gets \operatorname{MLP}(\mathbf{G})$ \Comment{project global features $\mathbf{G}$ to $\mathbb{R}^{d_k}$}
        \State $x \gets \operatorname{MLPMixer}([x_{m}, g])$ \Comment{pass regular and global tokens through the MLP-Mixer}
    \Else
        \State $x \gets \operatorname{MLPMixer}(x_m)$ \Comment{pass only the regular tokens through the MLP-Mixer}
    \EndIf

\ElsIf{SubGVP}
    \State $x_m^{\text{GVP}}, \mathbf{v}_m^{\text{GVP}} \gets \operatorname{GVP}(x_m^{\text{MP}}, \mathbf{v}_m^{\text{MP}})$ \Comment{Algorithm \ref{alg:gvp}}
    \If{subformer \textbf{or} submixer}
        \State $x \gets \text{subformer or submixer}(x_m^{\text{GVP}})$ \Comment{apply subformer or submixer block to GVP features}
    \EndIf
    
\EndIf
\State $x\gets \operatorname{MLP}(x)$
\State \Return $x$ \Comment{$x\in \mathbb{R}^{d_o}$}
\end{algorithmic}
\end{algorithm}
\end{minipage}
\end{figure*}

\clearpage
\section{Training details and hyperparameters}\label{appendix:training}

For pre-training, we use the Orbnet Denali dataset from Ref.\ \citenum{christensen2021orbnet}, which consists of 2.3 million conformations of small to medium size molecules. These conformations are obtained with semi-empirical methods and contain nonequilibrium geometries, alternative tautomers, and non-bonded interactions. For pretraining, we randomly take 10,000 configurations as the validation set and the remainder as the training set. 

\hspace*{-1em}
\begin{table}[htb]
    \centering
    \begin{minipage}[t]{0.45\textwidth}
        \centering
        \caption{Hyperparameters for pretrained networks}
        \vspace{0pt} 
        \begin{tabular}{c|c}\hline
            \textbf{Hyperparameter} & \textbf{TorchMD-ET and ViSNet} \\
            \hline
            $d$    & 64, 128, 256, 384 \\
            \# of MP layers     & 6 \\
            \# of attention heads & 8 \\
            Batch size          & 100 \\
            Epochs              & 10 \\
            \# of RBFs           & 64 \\
            $r_{\text{cut}}$ ($\text{\AA}$)     & 5, 7.5 \\
            Learning rate       & 0.0005 \\
            Optimizer           & AdamW (AMSGrad) \\
            Noise level ($\text{\AA}$)         & 0.2 \\ \hline
        \end{tabular}
        \label{tab:hp_pretrain}
    \end{minipage}%
    \hspace*{1em}
    \begin{minipage}[t]{0.45\textwidth}
        \centering
        \caption{Hyperparameters for SPIB}
        \vspace{0pt} 
        \begin{tabular}{c|cc}\hline
            \textbf{Hyperparameter}  & \multicolumn{2}{c}{\textbf{Shared}} \\ \hline
            Batch size & \multicolumn{2}{c}{1000} \\
            Learning rate & \multicolumn{2}{c}{0.0002} \\
            Optimizer & \multicolumn{2}{c}{AdamAtan2} \\
            Training patience & \multicolumn{2}{c}{5} \\
            MLP hidden dimension & \multicolumn{2}{c}{64} \\
            MLP activation function & \multicolumn{2}{c}{SiLU} \\
            Embedding dimension ($d$) & \multicolumn{2}{c}{64} \\
            Dropout & \multicolumn{2}{c}{0.2} \\
            Batch normalization & \multicolumn{2}{c}{False} \\ 
            Label refinement frequency & \multicolumn{2}{c}{5} \\
            Architectures & \multicolumn{2}{c}{SubFormer/SubMixer} \\
            GVP & \multicolumn{2}{c}{Yes} \\
            \# of GVP layers & \multicolumn{2}{c}{3} \\
            \# of transformer/mixer layers & \multicolumn{2}{c}{3} \\
            Expansion factor & \multicolumn{2}{c}{2} \\ 
            Global token & \multicolumn{2}{c}{True} \\ \hline
             & \textbf{Villin} & \textbf{Trp-cage} \\ \hline
            Trajectory stride & 2 & 4 \\
            Training lag time (ns) & 20 & 10 \\
            \hline
        \end{tabular}
        \label{tab:hp_spib}
    \end{minipage} 
\end{table}

\begin{table*}[h!]
\centering
\caption{Hyperparameters for VAMPnets. For GVP variants tested on villin, the architecture is consistent with the ones used in SPIB tasks.}
\begin{tabular}{c|ccc}\hline
     \textbf{Hyperparameter}  & \textbf{Chignolin} & \textbf{Villin} & \textbf{Trp-cage} \\ \hline
     Batch size & 5000 & 5000/1000&5000/1000\\
     Learning rate & 0.0002& 0.0002&0.0002\\
     Optimizer & AdamAtan2\cite{everett_scaling_2024} &AdamAtan2&AdamAtan2 \\
     Maximum epochs & 20& 20 &20 \\
     Training patience &5& 500& 500 \\
     Validation patience &2& 10& 10 \\
     Validation interval &5& 50 & 50 \\
     MLP hidden dimension & 256 & 128/64 & 128/64 \\
     MLP activation function &SiLU&SiLU&SiLU \\
     Trajectory stride &1&2&4\\
     Training lag time (ns) &4 &20&10\\
     Embedding dimension ($d$) &256&128/64&128/64\\
     Dropout &0.2&0.2&0.2\\
     Batch normalization &False&False&False\\
     Output dimension ($d_o$) &2&3&4 \\
     Architectures &MLP/SubFormer/SubMixer& MLP/SubFormer/SubMixer&MLP/SubFormer/SubMixer\\
     GVP & No & Yes & No \\
        Expansion factor & 2 & 2 & 2 \\ 
     Global token & False&True&False \\
     \hline
\end{tabular}
\label{tab:hp_vamp_pretrain}
\end{table*}

\newpage
\section{Supplementary Figures}
\begin{figure*}[!h]
    \centering
    \includegraphics[width=\textwidth]{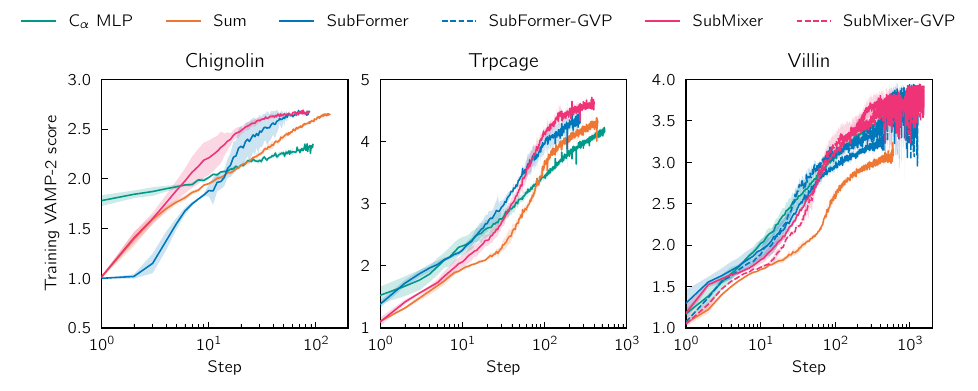}
    \caption{Training curves for VAMPnets. Each curve represents the mean of three training runs. Shaded regions indicate standard error over three runs.}
    \label{fig:train_curves}
\end{figure*}
\begin{figure*}[!h]
    \centering
    \includegraphics[width=\textwidth]{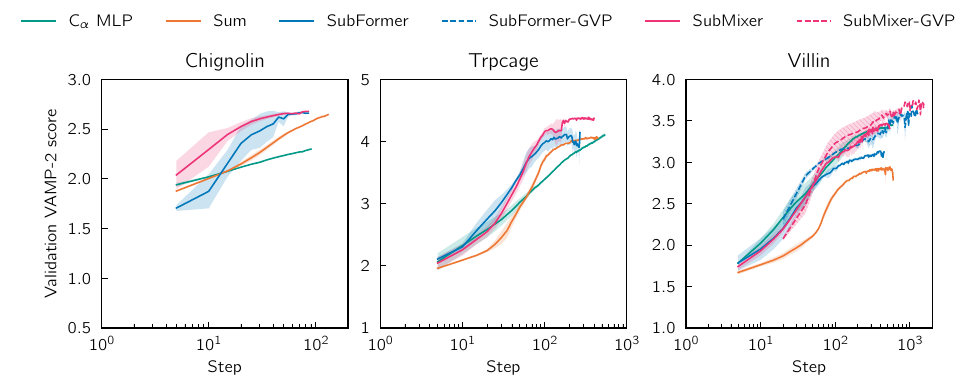}
    \caption{Validation curves for VAMPnets. Each curve represents the mean of three training runs. Shaded regions indicate standard error over three runs.}
    \label{fig:valid_curves}
\end{figure*}

\begin{figure*}[!h]
    \centering
    \includegraphics[width=\textwidth]{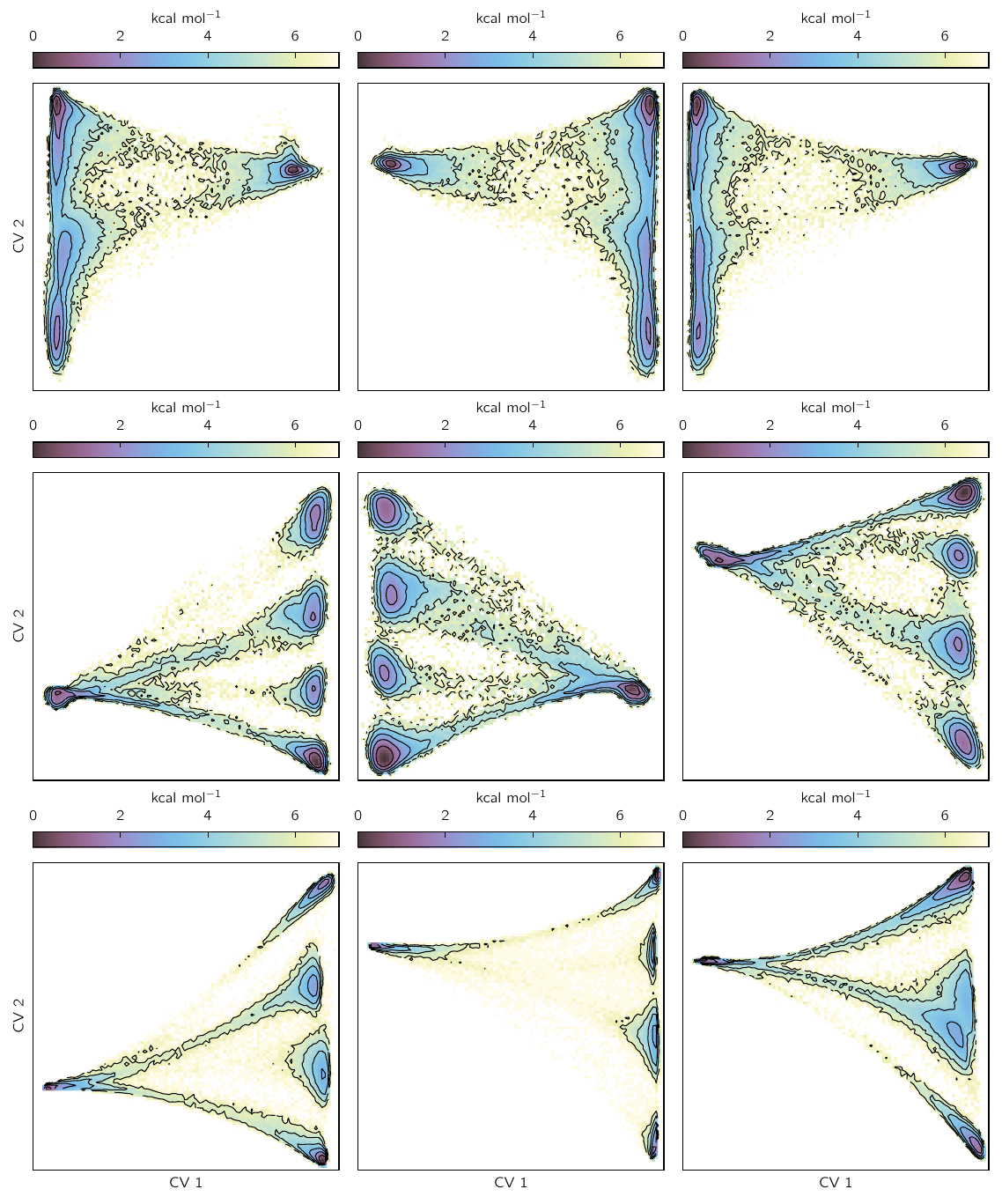}
    \caption{PMFs as a function of VAMP CVs for chignolin. From top to bottom, VAMPnets were trained with no token mixer (Sum), SubMixer, or SubFormer. Each column shows the result from a single training run. Contours are drawn every 1 kcal/mol.}
    \label{fig:vamp_chig_cvs}
\end{figure*}

\begin{figure*}[!h]
    \centering
    \includegraphics{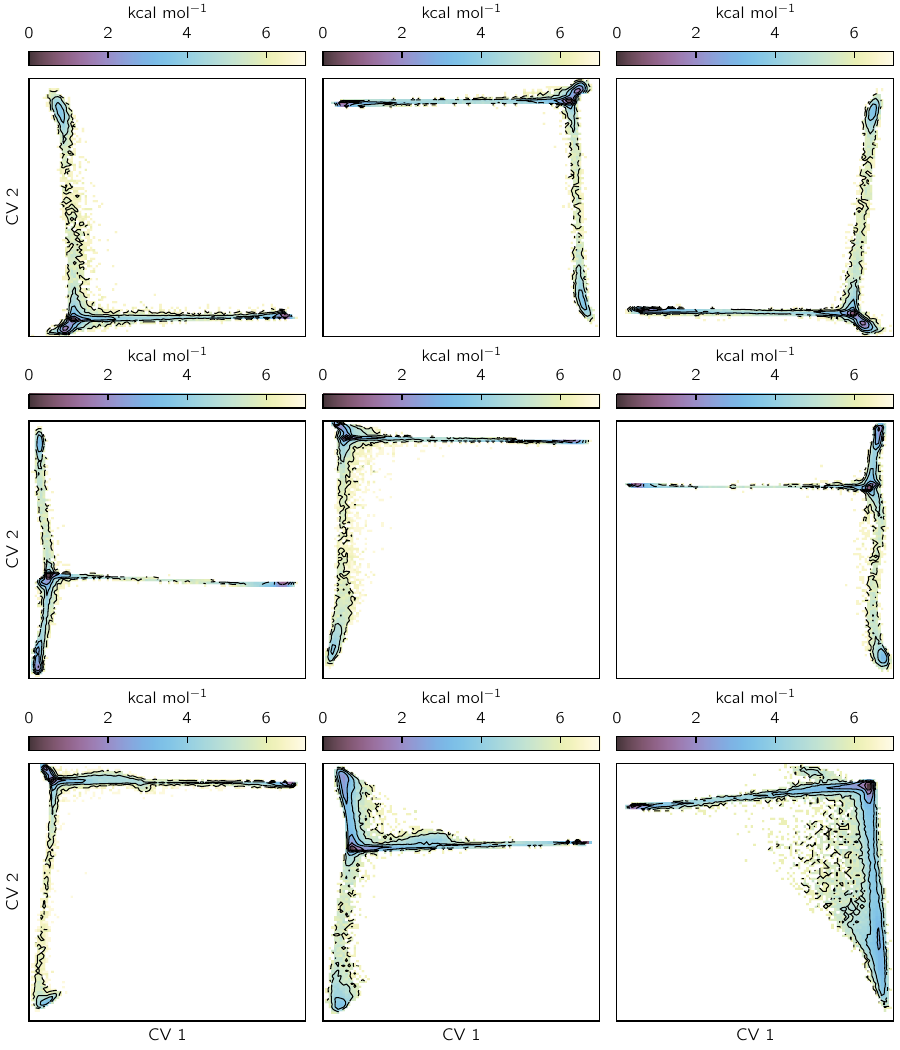}
    \caption{PMFs as a function of VAMP CVs for trp-cage. From top to bottom, VAMPnets were trained with no token mixer (Sum), SubMixer, or SubFormer. Each column shows the result from a single training run. Contours are drawn every 1 kcal/mol.}
    \label{fig:vamp_trpcage_cvs}
\end{figure*}

\begin{figure*}[!h]
    \centering
    \includegraphics[scale=0.95]{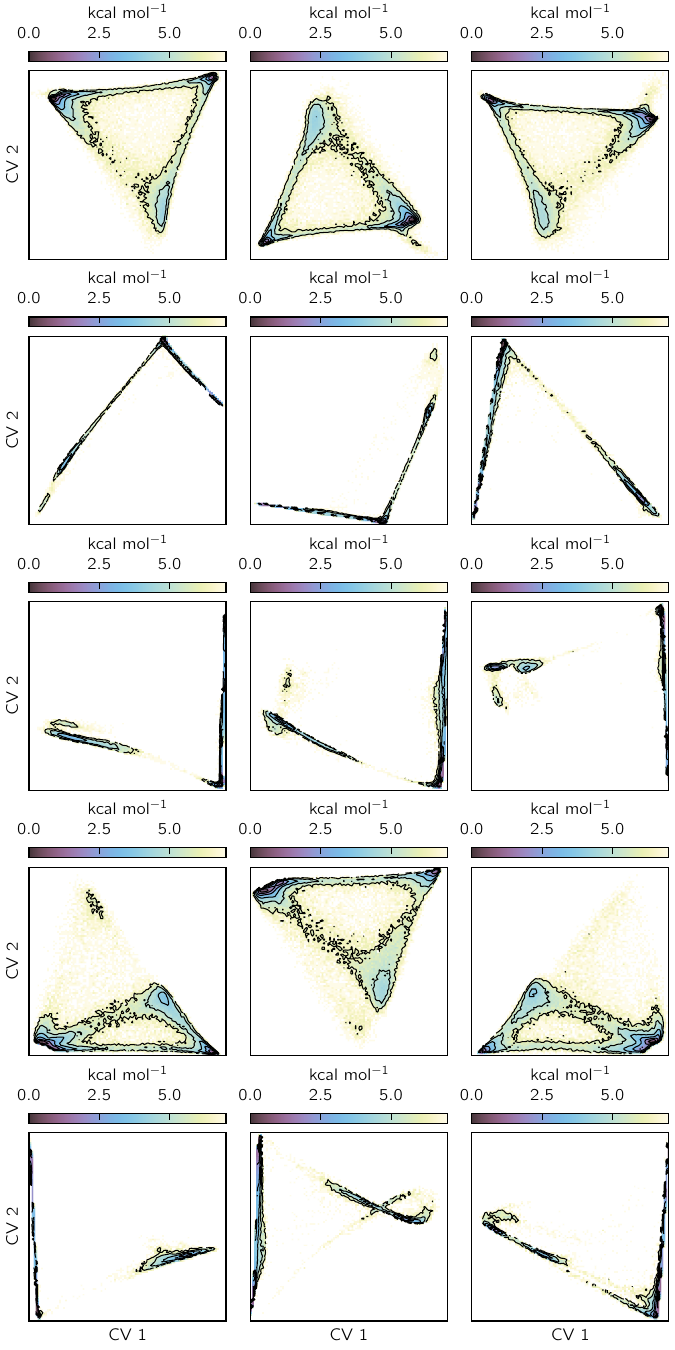}
    \caption{PMFs as a function of VAMP CVs for villin. From top to bottom, VAMPnets were trained with no token mixer (Sum), SubMixer, SubMixer-GVP, SubFormer, or SubFormer-GVP. Each column shows the result from a single training run. Contours are drawn every 1 kcal/mol.}
    \label{fig:vamp_villin_cvs}
\end{figure*}

\begin{figure*}[!h]
    \centering
    \includegraphics{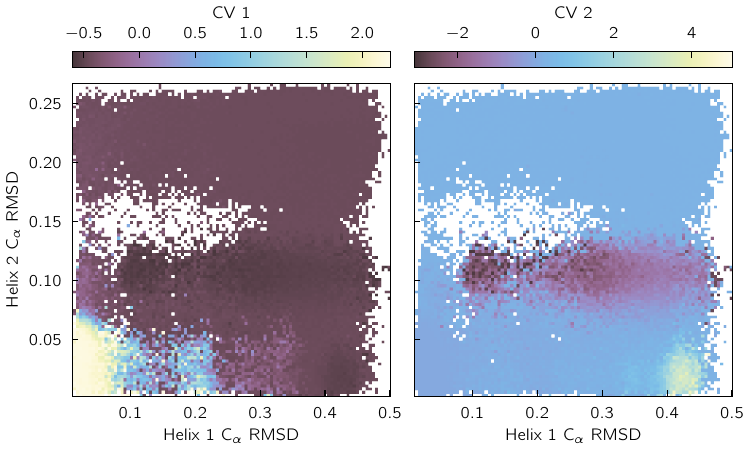}
    \caption{Trp-cage VAMPnet (with SubMixer) CVs as a function of two physical coordinates: C$_\alpha$ RMSD of helix 1 (residues 2--9) and C$_\alpha$ RMSD of helix 2 (residues 11--14). The C$_\alpha$ RMSDs were computed with respect to the PDB structure 2JOF \cite{barua_trp-cage_2008}.}
    \label{fig:vamp_trpcage_physical}
\end{figure*}

\begin{figure*}[!h]
    \centering
    \includegraphics{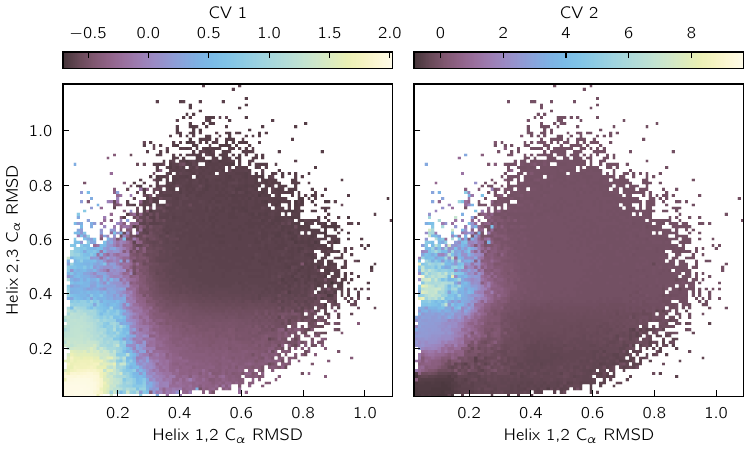}
    \caption{Villin VAMPnet (with SubFormer) CVs as a function of two physical coordinates: C$_\alpha$ RMSD of helices 1 and 2 (residues 3--10 and 14--19), and C$_\alpha$ RMSD of helices 2 and 3 (residues 14--19 and 22--32). The C$_\alpha$ RMSDs were computed with respect to the PDB structure 2F4K \cite{kubelka_sub-microsecond_2006}.}
    \label{fig:vamp_villin_physical}
\end{figure*}

\begin{figure*}[!h]
    \centering
\includegraphics[width=0.9\textwidth]{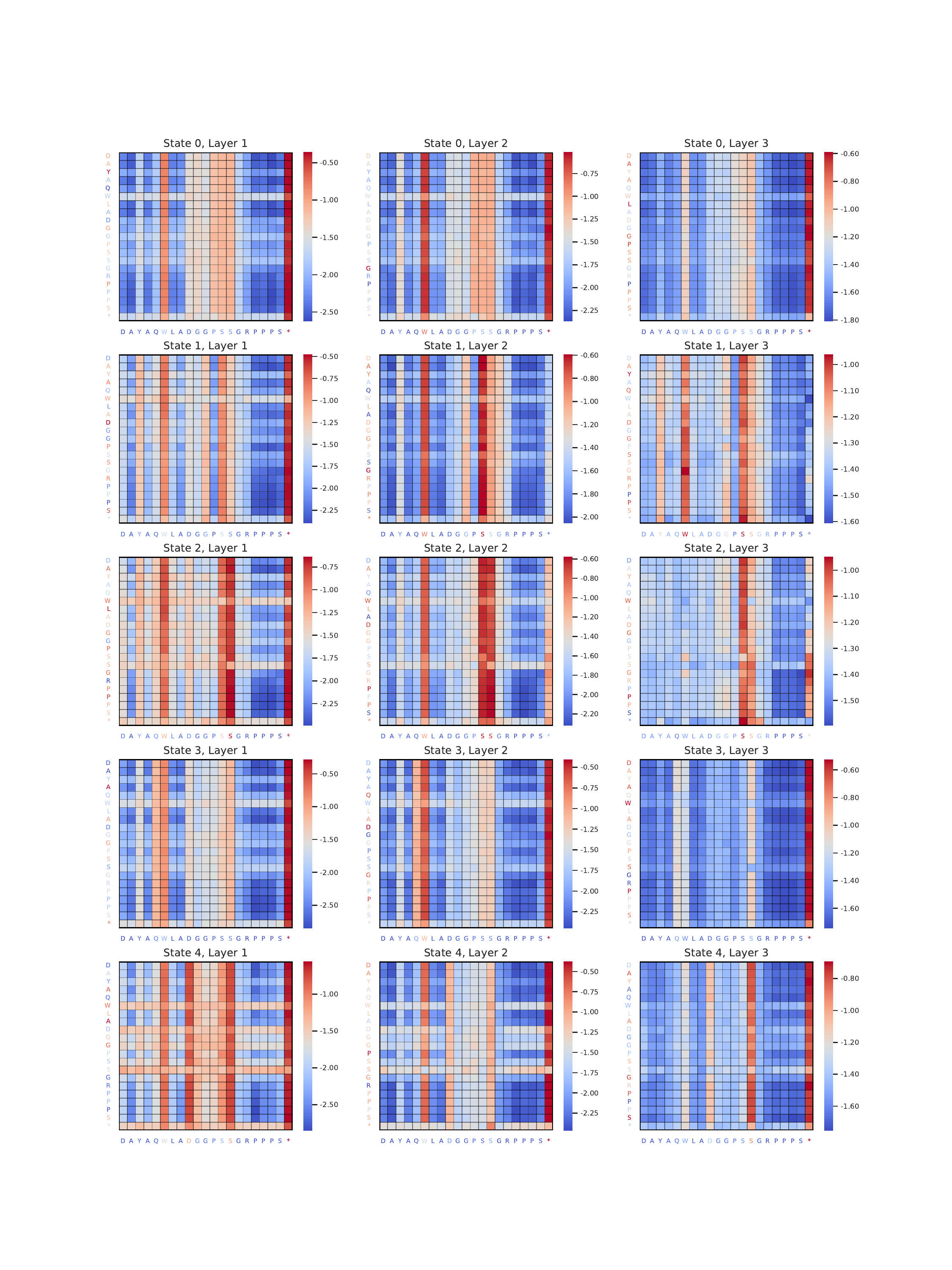}
    \caption{Log-scaled attention weight heatmaps for trp-cage SPIB states 0 to 4 from three layers of SubFormer-GVP. Each subplot displays attention weights with color-coded tick labels based on normalized sums. Colorbars indicate log-scaled attention values.}
    \label{fig:spib_attn_map_trpcage_0}
\end{figure*}

\begin{figure*}[!h]
    \centering
\includegraphics[width=0.9\textwidth]{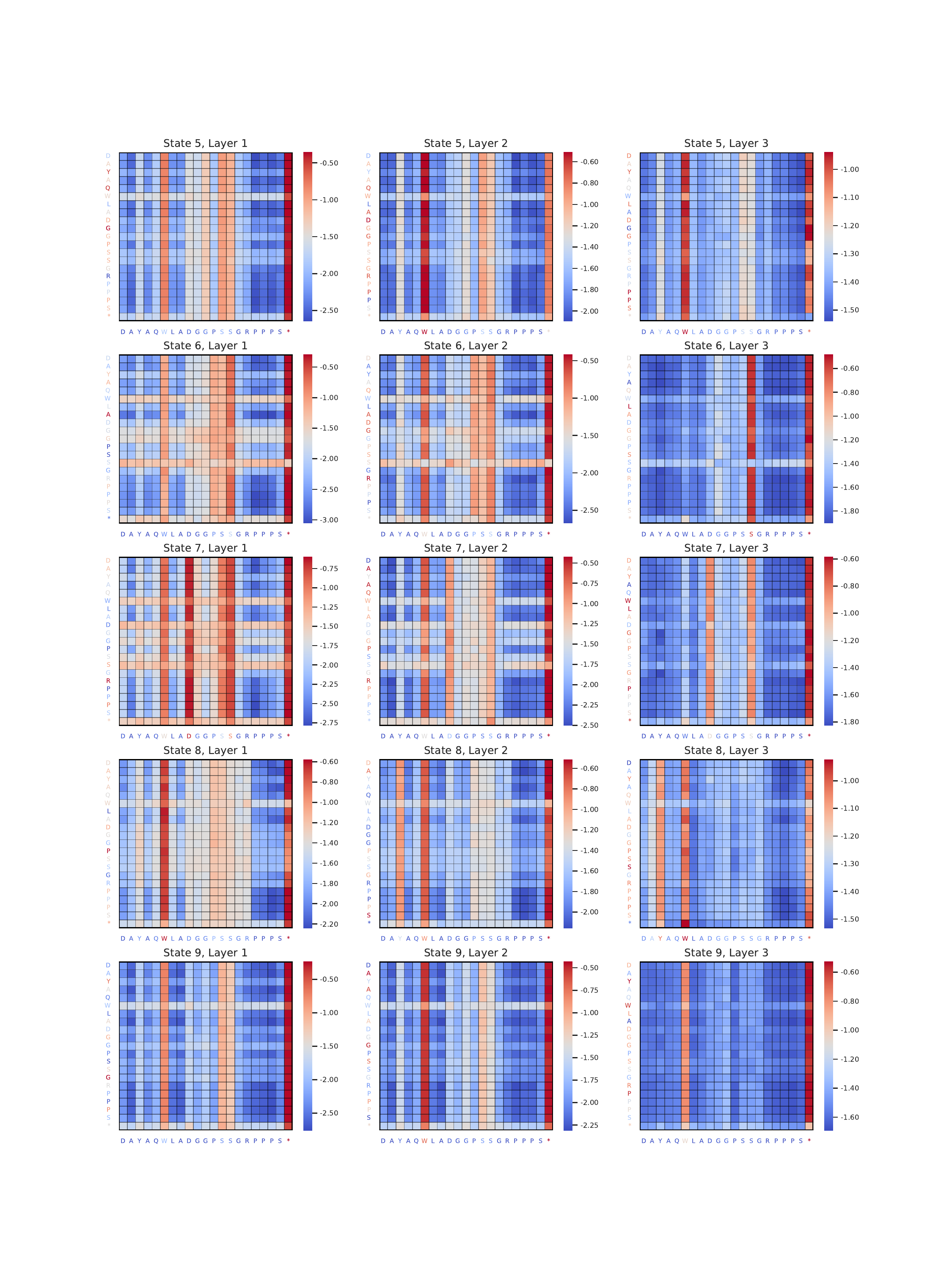}
    \caption{Log-scaled attention weight heatmaps for trp-cage SPIB states 5 to 9 from three layers of SubFormer-GVP. Each subplot displays attention weights with color-coded tick labels based on normalized sums. Colorbars indicate log-scaled attention values.}
    \label{fig:spib_attn_map_trpcage_1}
\end{figure*}

\begin{figure*}[!h]
    \centering
\includegraphics[width=\textwidth]{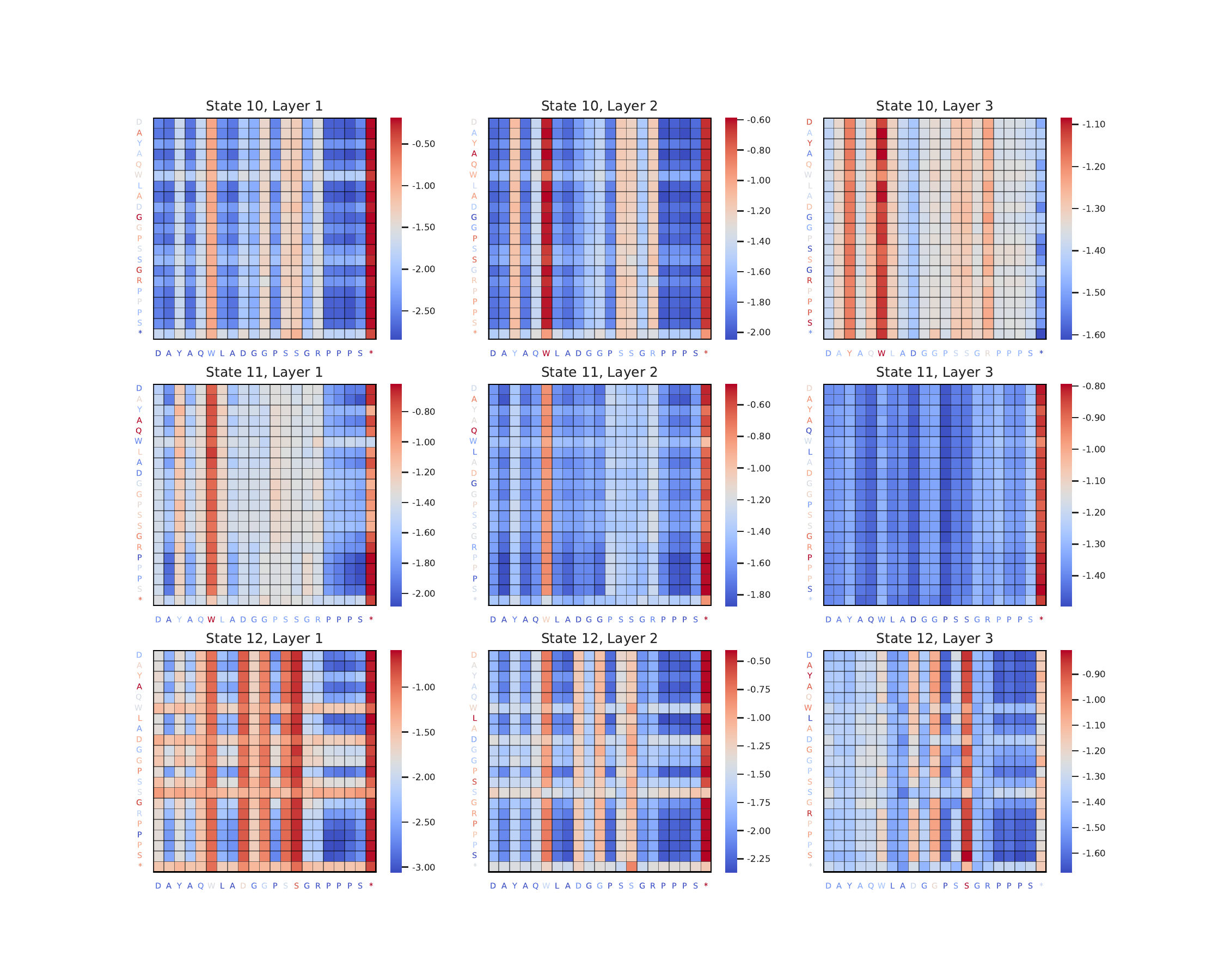}
    \caption{Log-scaled attention weight heatmaps for trp-cage SPIB states 10 to 12 from three layers of SubFormer-GVP. Each subplot displays attention weights with color-coded tick labels based on normalized sums. Colorbars indicate log-scaled attention values.}
    \label{fig:spib_attn_map_trpcage_2}
\end{figure*}

\begin{figure*}[!h]
    \centering
\includegraphics[width=\textwidth]{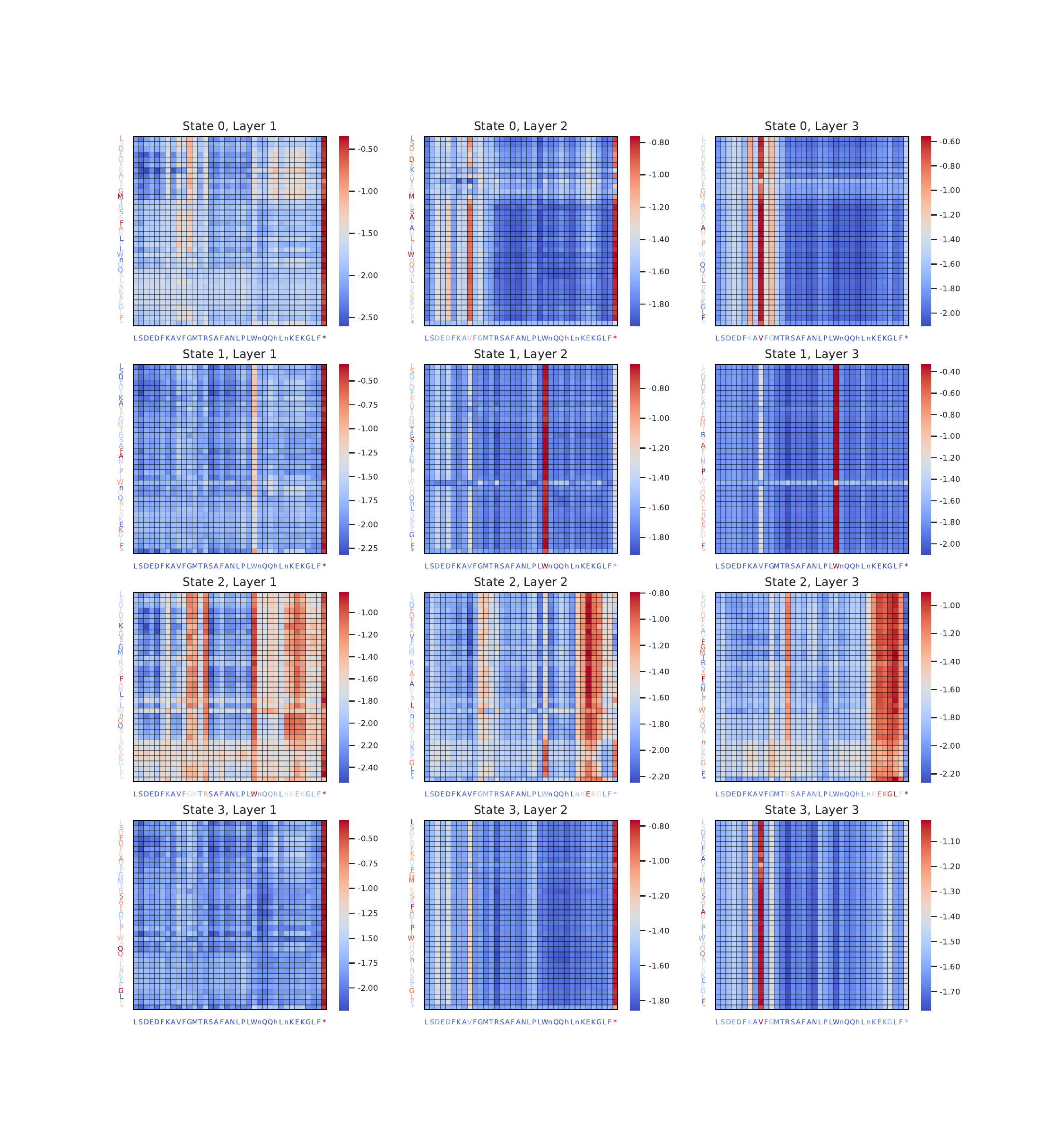}
    \caption{Log-scaled attention weight heatmaps for villin SPIB states 0 to 3 from three layers of SubFormer-GVP. Each subplot displays attention weights with color-coded tick labels based on normalized sums. Colorbars indicate log-scaled attention values.}
    \label{fig:spib_attn_map_villin_0}
\end{figure*}

\begin{figure*}[!h]
    \centering
\includegraphics[width=\textwidth]{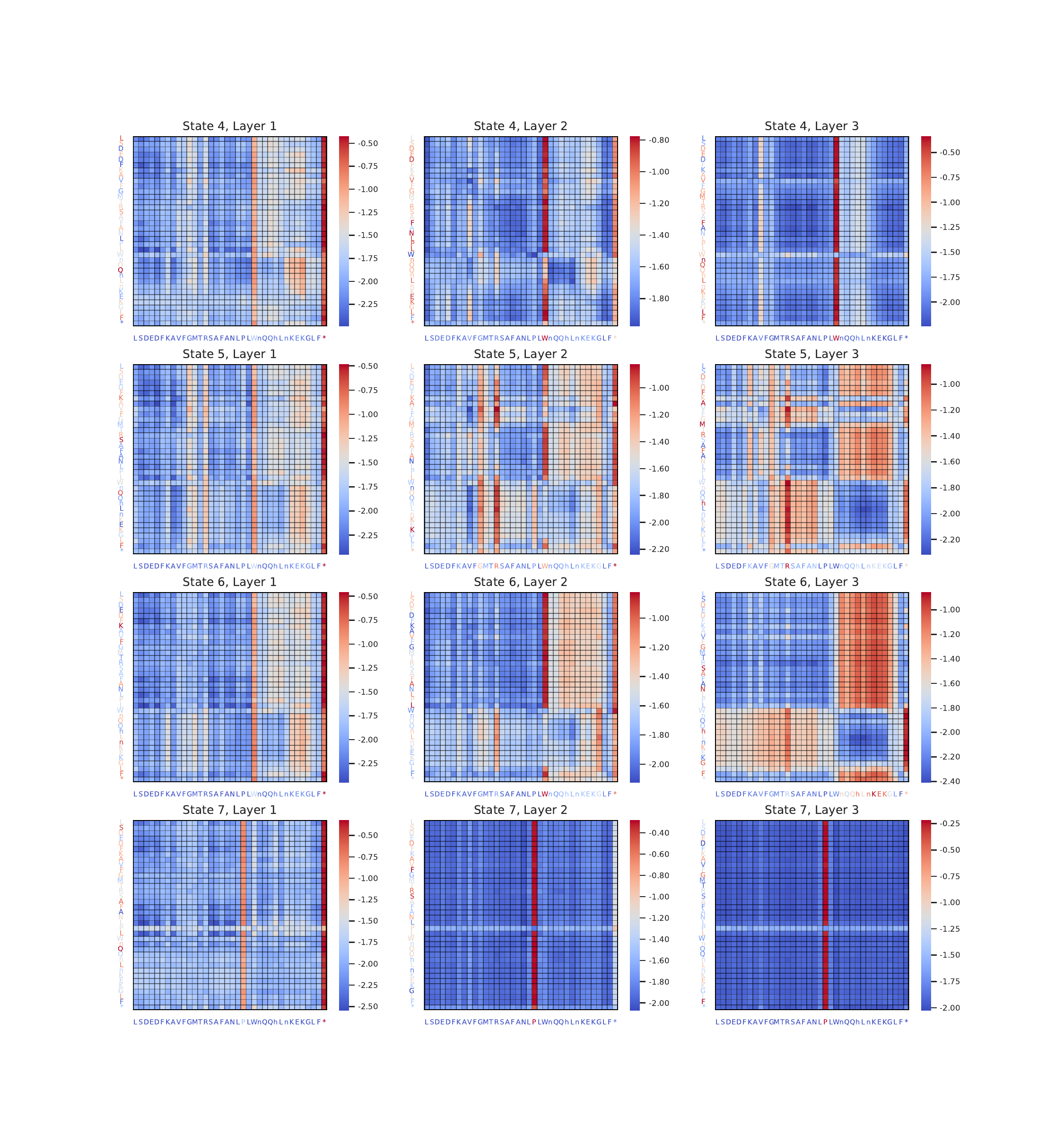}
    \caption{Log-scaled attention weight heatmaps for villin SPIB states 4 to 7 from three layers of SubFormer-GVP. Each subplot displays attention weights with color-coded tick labels based on normalized sums. Colorbars indicate log-scaled attention values.}
    \label{fig:spib_attn_map_villin_1}
\end{figure*}

\begin{figure*}[!h]
    \centering
\includegraphics[width=\textwidth]{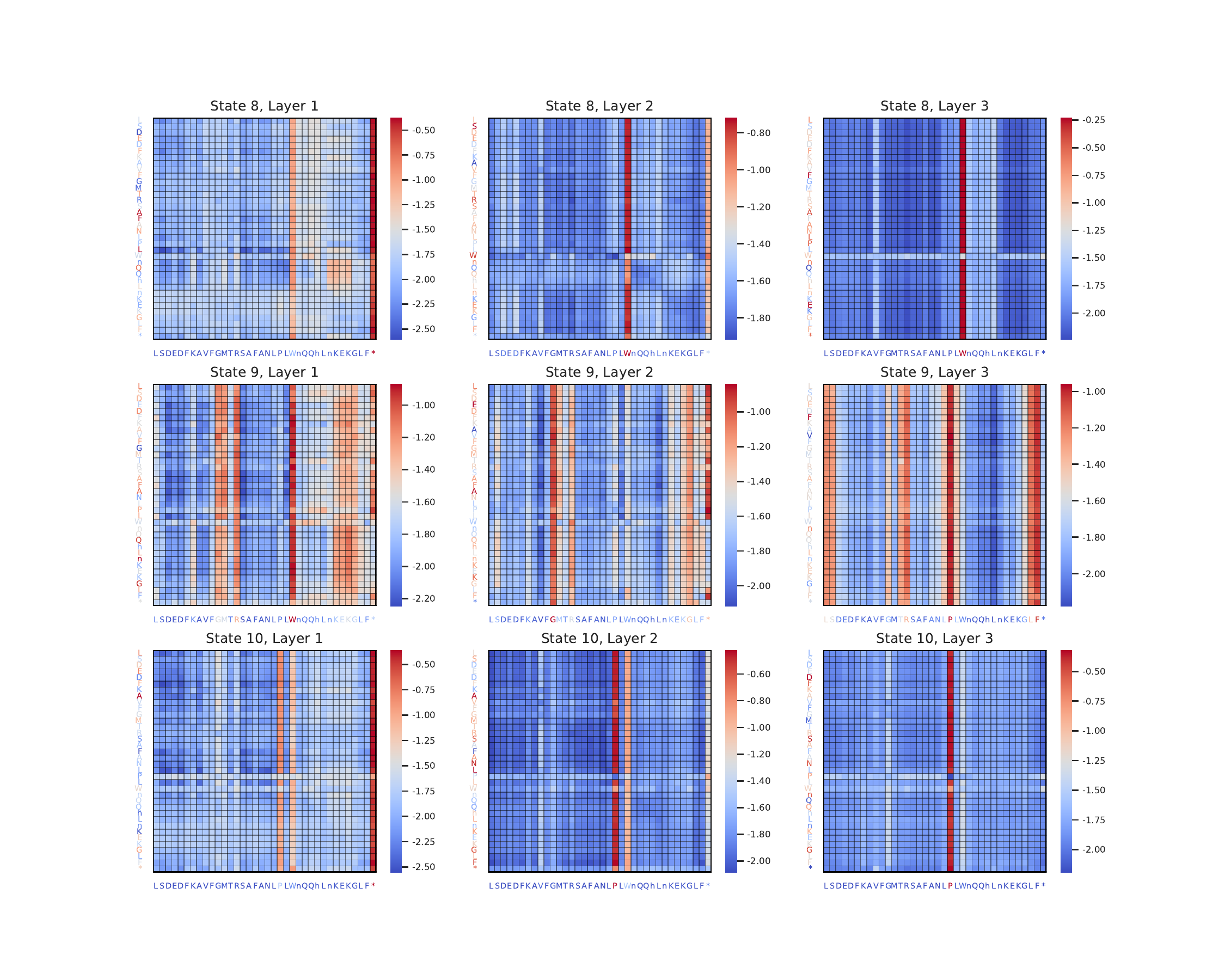}
    \caption{Log-scaled attention weight heatmaps for villin SPIB states 8 to 10 from three layers of SubFormer-GVP. Each subplot displays attention weights with color-coded tick labels based on normalized sums. Colorbars indicate log-scaled attention values.}
    \label{fig:spib_attn_map_villin_2}
\end{figure*}

\begin{figure*}[!h]
    \includegraphics[width=\textwidth]{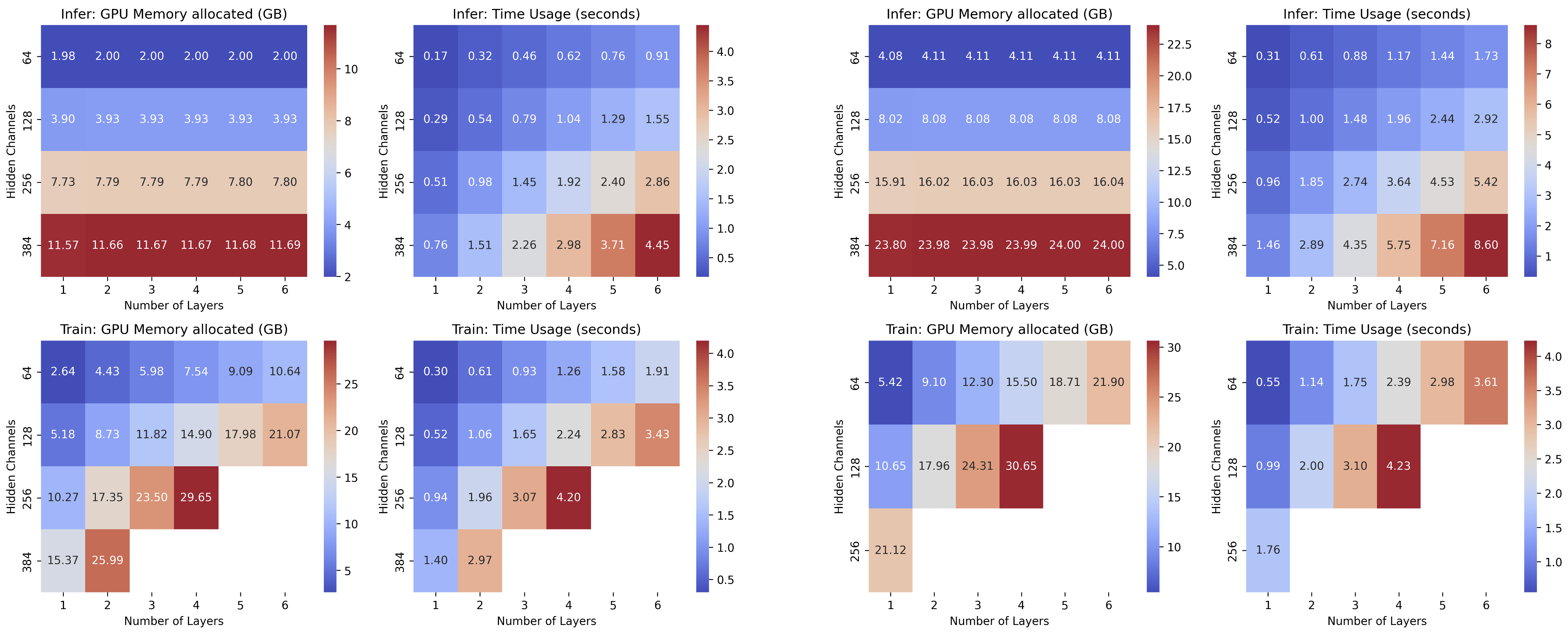}
    \caption{Comparison of time and GPU memory usage of TorchMD-ET in training and inference modes on the trp-cage (left) and villin (right) subsets, each consisting of 2000 frames with 100 frames per batch. The measurements are done without specific objective functions for usage benchmarking purposes. The model's performance is evaluated across varying numbers of hidden channels (64, 128, 256, 384) and layers (1--6). The heatmaps on the left show GPU memory allocated in gigabytes (GB), while the heatmaps on the right depict time usage in seconds. The time represents one forward pass for inference mode and one forward plus one backward pass for training mode. Measurements were performed on an NVIDIA A100 GPU with 40G of memory. Missing values are due to out-of-memory errors for certain configurations.}
    \label{fig:combined_usage}
\end{figure*}

\end{document}